\documentclass[letterpaper, 10 pt, conference]{ieeeconf}  
\IEEEoverridecommandlockouts                              
\usepackage{algorithm}
\usepackage[noend]{algpseudocode}
\algrenewcommand\algorithmicindent{0.7em}%

\usepackage{mathtools}

\usepackage[dvipsnames]{xcolor}
\usepackage{cellspace}
\usepackage{glossaries-extra}
\setabbreviationstyle[acronym]{long-short}

\usepackage{adjustbox}
\usepackage{cite}

\usepackage{bm}
\usepackage{breqn}
\DeclareMathOperator*{\argmax}{arg\,max}

\usepackage{amsthm} 
\usepackage{amsfonts}
\usepackage{amsmath}
\newtheorem{problem}{Problem}

\theoremstyle{definition}

\usepackage{graphicx}

\title{\LARGE \bf
Decentralised Active Perception in Continuous Action Spaces\\ for the Coordinated Escort Problem
} 

\author{Rhett Hull$^{1,3}$, Ki Myung Brian Lee$^1$, Jennifer Wakulicz$^1$, Chanyeol Yoo$^1$, James McMahon$^2$, Bryan Clarke$^3$, \\Stuart Anstee$^3$, Jijoong Kim$^3$ and Robert Fitch$^{1}$
\thanks{This research is supported by Australian Government Research Training Program (RTP) Scholarships, the Commonwealth of Australia and the Office of Naval Research (N62909-21-1-2031).}
\thanks{$^1$Authors are with the University of Technology Sydney, Ultimo, NSW 2006, Australia {\tt\footnotesize \{rhett.c.hull, brian.lee, jennifer.wakulicz\}@student.uts.edu.au, \{chanyeol.yoo, rfitch\}@uts.edu.au}.}
\thanks{$^{2}$James McMahon is with the US Naval Research Laboratory, USA
        {\tt\footnotesize james.mcmahon@nrl.navy.mil}.}%
\thanks{$^{3}$Authors are with the Defence Science and Technology Group, Department of Defence, Australia
        {\tt\footnotesize \{rhett.hull1, bryan.clarke, stuart.anstee, jijoong.kim\}@defence.gov.au}.}%
}

\begin{document}

\maketitle
\thispagestyle{empty}
\pagestyle{empty}

\begin{abstract}
We consider the \emph{coordinated escort problem}, where a decentralised team of supporting robots implicitly assist the mission of higher-value principal robots.
The defining challenge is how to evaluate the effect of supporting robots' actions on the principal robots' mission.
To capture this effect, we define two novel auxiliary reward functions for supporting robots called satisfaction improvement and satisfaction entropy, which computes the improvement in probability of mission success, or the uncertainty thereof.
Given these reward functions, we coordinate the entire team of principal and supporting robots using decentralised cross entropy method (Dec-CEM), a new extension of CEM to multi-agent systems based on the product distribution approximation. 
In a simulated object avoidance scenario, our planning framework demonstrates up to two-fold improvement in task satisfaction against conventional decoupled information gathering.
The significance of our results is to introduce a new family of algorithmic problems that will enable important new practical applications of heterogeneous multi-robot systems.

\end{abstract}


\section{INTRODUCTION}
Applications of coordinated multi-robot systems can involve heterogeneous teams where a principal robot is assisted in some way by one or more supporting robots, which may be less capable and of lower cost. Supporting robots can play various roles, such as acting as a source of remote sensing and perception to better inform navigation decisions. One of the fundamental challenges in designing such systems is how to coordinate the behaviour of supporting robots to facilitate the progress of the principal robot in achieving its goal. It is desirable for supporting robots to actively collect information that is relevant to the principal robot's goals, but the actions of the principal robot in turn may depend on such information and thus are not known in advance.

Coordination algorithms that would enable a team of robots to support a principal robot or agent must be able to predict the effect that the supporting team's measurements or actions have on the principal agent's task performance. We introduce the term \emph{coordinated escort problem} to refer to this class of coordination problems, in the sense that supporting robots act as an escort team for the principal robot/agent.

Escorting roles are prevalent during conflicts and emergencies; for example, when ships or vehicles must pass through an area that is suspected of being mined, and when convoys must pass along routes that are open to attack. An implicit aspect of this process is that escorts accept increased risk of harm on behalf of those they are escorting; this drives research into the use of robots in such roles. There is typically an assumption in such cases that the escorts will act somewhat independently of those they are protecting. Escorts cannot assume that their behaviours will be coordinated, or even understood, by the principal agent. On the other hand, escorts must understand the goals of the principal agent to behave appropriately. Escorts must also be capable of communicating information about risk and safety, because it is likely that the principal agent will have to modify its own behaviour to make the escort's task feasible. 

\begin{figure}[t!]
    \centering
    \includegraphics[width=1\linewidth]{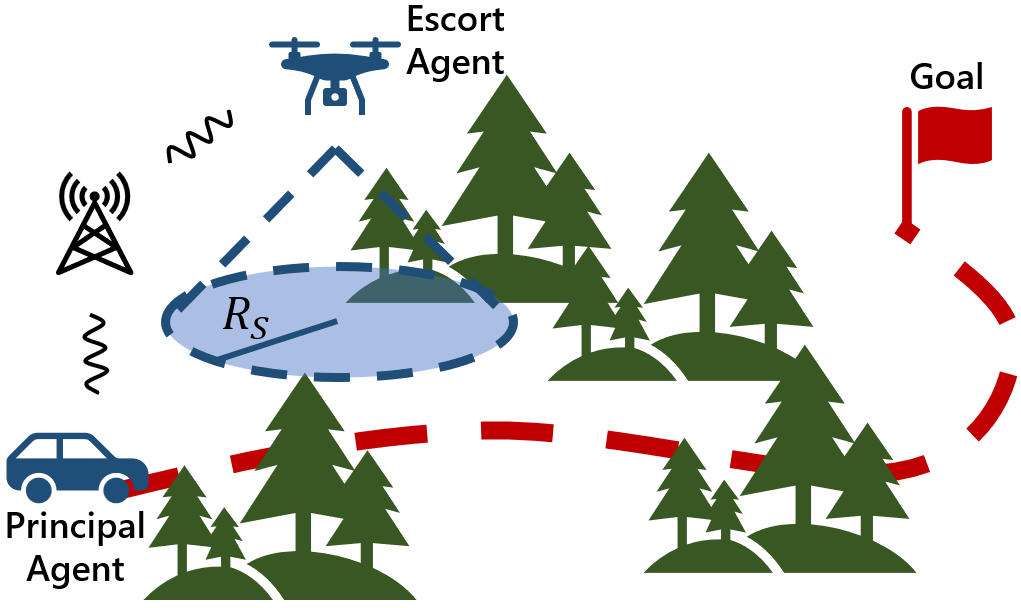}
    \vspace{-3ex}
    \caption{Example of coordinated escort. The principal agent with no on-board sensing is tasked with avoiding the objects and reaching the goal. The escort agent is equipped with a sensor to take measurements and update the belief over object locations. Measurements and trajectory intent are communicated between all agents. We consider escort teams with up to three escort agents.}
    \vspace{-3ex}
    \label{fig:prob_statement_diagram}
\end{figure}


In this paper, we define a specific instance of the general coordinated escort problem and present DecCEM, a decentralised solution based on a novel variant of the cross-entropy method~(CEM) for planning with continuous actions. The problem we consider is where escort agents (EA) must perform information gathering with uncertain object locations modelled by Gaussian beliefs, in order to support a principal agent (PA). 
CEM~\cite{cem_kroese2006,cem_for_planning} is a sampling-based planning algorithm that finds a control distribution that probabilistically maximises a given reward. 
CEM admits arbitrary parameterisation of the control distribution, and thus extends product distribution-based planners~\cite{best2019decmcts,wolpert} from discrete to continuous action spaces. Our approach in developing DecCEM is to extend CEM to multi-agent systems based on the product distribution approximation~\cite{best2019decmcts,wolpert}.

DecCEM acts to maximise a joint reward function across the entire team, including both EAs and PA. To faithfully capture the effect of the EAs' measurements on the PA's task performance, we define an auxiliary reward function for the EAs. We propose two alternative auxiliary functions, satisfaction entropy~(SE) and satisfaction improvement~(SI), that directly measure improvement in the PA's probability of task satisfaction, or uncertainty of task success.

In an example reach-avoid scenario depicted in Fig.~\ref{fig:prob_statement_diagram}, we compare our algorithm to traditional approaches that do not consider the PA's actions online. 
The SE approaches demonstrates up to two-fold improvement over basic information gathering and three-fold over when a principal agent planning over the prior alone.
The SI approach performs up to twice as better as planning over the prior alone, and, interestingly, equally or worse than basic information gathering.   
Finally, we demonstrate that our objective functions adaptively adjust the exploration and exploitation of the prior to influence the trajectory of the PA to improve overall mission success.

The contribution of this work is: 1) to introduce the coordinated escort problem as a specific type of joint optimisation problem where the coordination objective depends on the actions of an independent agent, and 2) to introduce the DecCEM algorithm that solves this problem for decentralised agents with continuous action spaces. This work helps to enable important practical applications of multi-robot systems where robots play an essential supporting role.







%

\section{RELATED WORK}


There are a range of interpretations of the escorting problem. Early variants define an optimal control problem where escort agents must entrap and shepherd principal robots to a goal region~\cite{distctrl_lan2010, shepherding_lien2004, distopt_monti2014, globalref_monti2014, shepherding_strombom2014, ee_antonelli2007, ee_antonelli2008, ee_mas2009}. More complex definitions come recently, in the form of \emph{coordinated} escort problems, where control for escorting and principal agents are jointly addressed to achieve a goal task~\cite{Wu_coord_2021, TASE_coord_2019}. Coordinated escort formulations of particular relevance focus heavily on the escort agent solving an information gathering problem to aid in goal completion~\cite{mars_nilsson2018, mars_sasaki2020, mars_folsom2021, brian2021}. 

Typical information gathering approaches such as~\cite{atanasov2014, atanasov2015, schlotfeldt2018, jen2021, krause2007near, sensorschedACC2018, activeprcpt_ghasemi2019, sensorsched2010,chen2020broadcast,to2021estimation} employ sensors (mobile or otherwise) to take measurements to reduce the uncertainty of state estimates of any target system in the environment. Others focus instead on reducing uncertainty about environmental processes or phenomenon~\cite{lee2018active, lee2019online, jen2022}. In both cases uncertainty reduction is often achieved by assuming a Gaussian belief over the target or environment, which is maintained using variants of the Kalman filter or Gaussian Processes. 
In these works, information gain can be directly evaluated because the covariance update procedures used are independent of the explicit value of measurements.
We follow a similar formulation for our escort agent's information gathering mission to describe an uncertain environment and to evaluate the effect of information gathered by the escorts. In our case, however, the information gathering objective for our escort agent differs to these papers.

Similar to~\cite{mars_nilsson2018, mars_sasaki2020, mars_folsom2021, brian2021}, our proposed information gathering objectives for escort agents focus not on reducing environmental uncertainty overall, but on reducing uncertainty specifically to aid the principal robot in achieving its task. While these works rely on sequential planning for EAs and PAs to solve the joint control problem in a centralised manner, our work presents a fully decentralised solution inspired by concepts from~\cite{best2019decmcts}.

\section{PROBLEM FORMULATION}\label{sec:prob_form}
\begin{figure}[t]
    \centering
    \includegraphics[width=1\linewidth]{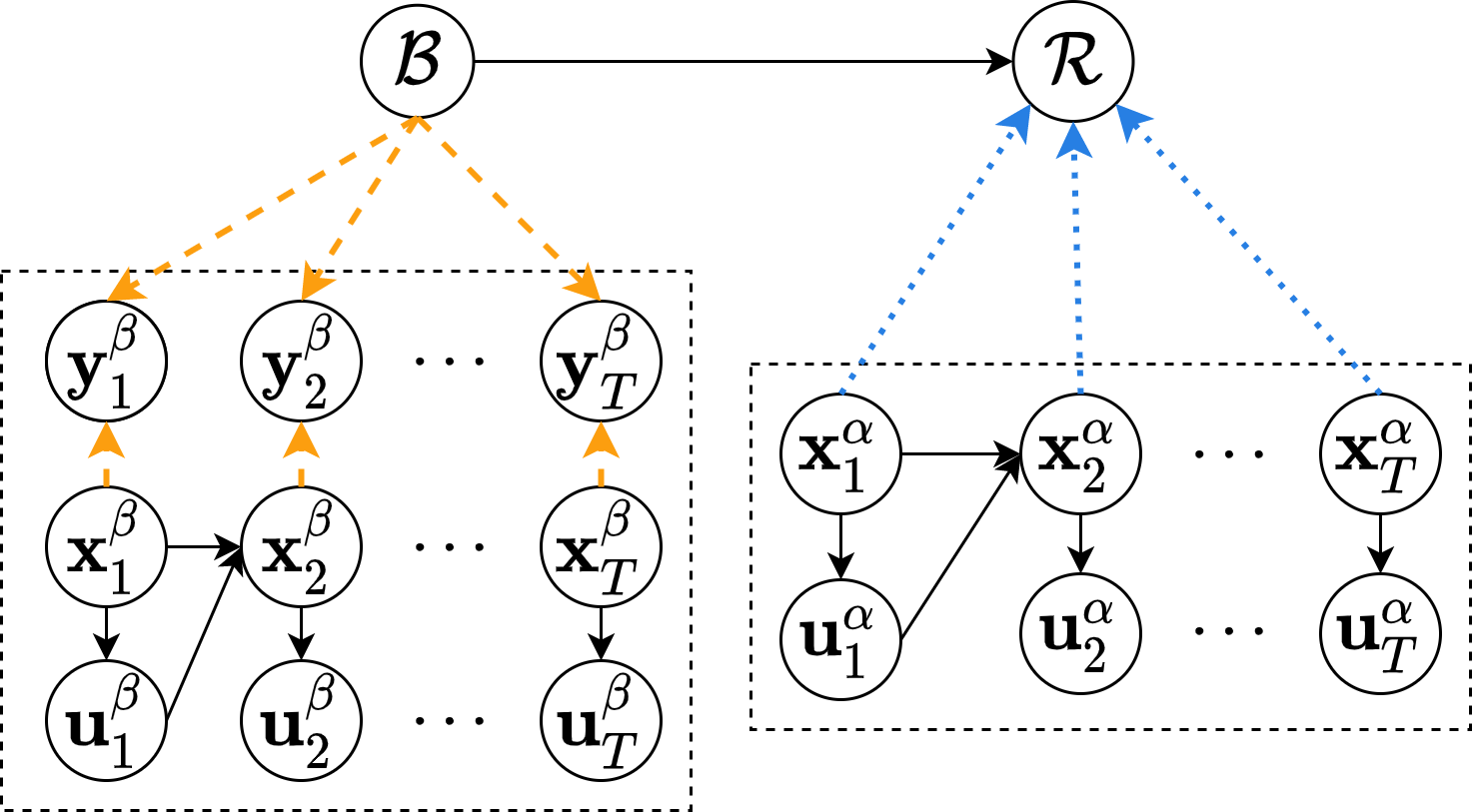}
    \caption{Probabilistic graphical model (PGM) representation of the decoupled escorting problem \cite{brian2021}. The escorting agent (blue dashed line) surveys the environment and measures $\mathbf{O}$. The principal agent (orange dotted line) moves to accomplish $\phi$ and may collide with $\mathbf{O}$. Collision results in a failed mission.}
    \vspace{-2ex}
    \label{fig:prob_form_diagram}
\end{figure}

The coordinated escort problem depicted in Fig.~\ref{fig:prob_form_diagram} comprises a team of robots $\mathcal{R}$.
The state $\mathbf{x}_{t}^{r}$ of each robot~$r \in \mathcal{R}$ at discrete time $t+1$ is described by a state transition model:
\begin{align} \label{eqn:dyn_model}
    \mathbf{x}_{t+1}^r &= \mathbf{f}^r(\mathbf{x}_t^r, \mathbf{u}_t^r)
    ,
\end{align}
where the states~$\mathbf{x}_{t}^{r} \in \mathbb{R}^{N}$ actions~$\mathbf{u}_{t}^{r} \in \mathbb{R}^{M}$ are continuous. 
The robots operate in a partially known environment comprising a set of objects, whose locations~$\mathbf{O} = \{\mathbf{o}_1, \cdots, \mathbf{o}_{N}\} \subset \mathbb{R}^{N}$ are known imprecisely. 

There are two distinct classes of robots, \emph{principal agents}~(PAs), and \emph{escort agents}~(EAs).
The set of PAs and EAs are disjoint subsets of $\mathcal{R}$ denoted by~$A,B \subset \mathcal{R}$ respectively.

Each EA $\beta \in B$ is equipped with a noisy sensor that can measure the locations of objects that are within a certain radial range~$R_S$.
Thus, the measurements~$\mathbf{y}_t^{\beta}$ at state~$\mathbf{x}_t^{\beta}$ are described by the sensor model:
\begin{equation}\label{eq:sensor_model}
    \mathbf{y}_{t}^{\beta}(\mathbf{x}_{t}^{\beta}) = \{\mathbf{o}_{i} + \epsilon \mid \mathbf{o}_i \in  \mathbf{O}, ||\mathbf{o}_i - \mathbf{x}_{t}^{\beta}|| < R_S\},
\end{equation}
where $\epsilon \sim \mathcal{N}(0,\sigma^{2} I)$ is zero-mean Gaussian noise.
We assume that all measurements $\mathbf{y}_{t}^{\beta}$ taken by each EA are made available to all other robots.
PAs are not equipped with sensors to take measurements, and must rely on the measurements from EAs to estimate the environment. 

For brevity, we write $\mathbf{X}_{t}^{r} = \{ \mathbf{x}^{r}_{t}, ... \mathbf{x}^{r}_{t+T} \}$ to mean the trajectory of robot $r \in \mathcal{R}$ over a time horizon $T$ starting from $t$.
The same applies to control actions and measurements (i.e., $\mathbf{U}_{t}^{r} = \{ \mathbf{u}^{r}_{t}, ... \mathbf{u}^{r}_{t+T} \}$, for $r \in \mathcal{R}$, $\mathbf{Y}_{t}^{\beta} = \{ \mathbf{y}^{\beta}_{t}, ... \mathbf{y}^{\beta}_{t+T} \}$, $\beta \in \mathcal{B}$).
Similarly, we replace the superscript with a set of robots to mean a set of trajectories of all robots in the set, e.g. $\mathbf{X}_{t}^{\mathcal{R}} = \{\mathbf{X}_{t}^{r} \mid r \in \mathcal{R} \}$.

Although the PAs cannot measure the object locations, they are required to complete a task $\phi$ that depends on the object locations $\mathbf{O}$.
The task is modelled in terms of satisfaction likelihood $P(\phi \mid \mathbf{X}^{A}_{t}, \mathbf{O})$, which describes the probability of success given \emph{fully known} object locations. 
This can be, for example, a Boltzman distribution $\log P(\phi \mid \mathbf{X}^{A}_{t}, \mathbf{O}) \propto -C(\mathbf{X}^{A}_{t}, \mathbf{O})$ for a generic cost function $C$~\cite{levine2018reinforcement}, or the probability of satisfaction of some temporal logic formula $\Psi$ so that $P(\phi \mid \mathbf{X}^{A}_{t}, \mathbf{O}) = P(\mathbf{X}^{A}_{t} \models \Psi \mid \mathbf{O})$~\cite{rstl_brian2021,yoo2015control,yoo2013provably,yoo2016online}.

Since the environment $\mathbf{O}$ is uncertain, $P(\phi \ |\ \mathbf{X}_t^{A}, \mathbf{O})$ cannot be directly computed.
Instead, we maximise the posterior probability of satisfaction $P(\phi \ | \ \mathbf{X}_t^{A}, \mathbf{Y}_{t}^{B})$ conditioned on the EAs' measurements $\mathbf{Y}_{t}^{B}$.
Doing so averages over all possible object locations according to the environmental belief afforded by the EA.
Therefore, the posterior probability of satisfaction captures the effect of measurements taken by EAs, and hence the quality of their paths. 
Overall, the problem is formally stated as follows.

\begin{problem}[Coordinated escort problem]\label{prob:coupled_escort}
    Given PAs~$A$, EAs~$B$ in a partially known environment~$\mathcal{E}$ with hidden objects~$\mathbf{O} \subset \mathcal{E}$, find optimal sequences of controls~$\mathbf{U}^{A*}$, ~$\mathbf{U}^{B *}$ for PAs and EAs respectively that maximises the PAs' probability of satisfying a task~$\phi$:
    \begin{equation}
        \mathbf{U}^{A*}, \mathbf{U}^{B*} = \argmax_{\mathbf{U}^{A}, \mathbf{U}^{B}}  P (\phi \mid \mathbf{X}_t^{A}, \mathbf{Y}_t^{B} ),
    \end{equation}
\end{problem}
The challenge of Problem~\ref{prob:coupled_escort} is twofold.
Most immediately, a naive approach would necessitate centralised planning due to the complicated dependence of the objective on the EAs' trajectory. 
We resolve this issue using a decentralised variant of the cross-entropy method~(CEM)~\cite{cem_kroese2006}, which can compute high-quality solutions as long as the objective can be evaluated given the robots' trajectories. 
More significantly, the values of future measurements $\mathbf{Y}_{t}^{B}$ are not available at planning time. 
We mitigate this issue by deriving alternative reward functions for EA that capture improvement in PA's planning performance with predicted Gaussian beliefs.

\section{PLANNING FOR COORDINATED ESCORT} \label{sec:contrib}
We propose a receding-horizon planning framework for solving Problem~\ref{prob:coupled_escort} that addresses the aforementioned challenges of decentralisation and measurement selection. 
Alg.~\ref{alg:overview} outlines the framework for a single robot $r$.
In line~\ref{alg:overview:belief}, the belief $\mathcal{B}_{t}(\mathbf{O})$ over object locations is updated with latest measurements. 
The updated belief is then used for evaluation of reward functions $R^{r}$ for which to optimise the controls (lines 2-6).
EAs and PAs solve are given different rewards $R^{r}$ to circumvent the challenge of measurement simulation.

To find controls that maximise the reward, we introduce DecCEM outlined in Alg.~\ref{alg:cem}.
Instead of a single set of controls, DecCEM finds a \emph{control distribution} $\mathcal{Q}(\mathbf{U}_{t}^{\mathcal{R}})$ that maximises the task satisfaction probability~$\mathcal{P}(\phi \mid \mathbf{X}_{t}^{A}, \mathbf{Y}_{t}^{B})$.
Inspired by~\cite{best2019decmcts,wolpert}, we impose a product distribution factorisation $\mathcal{Q}(\mathbf{U}_{t}^{\mathcal{R}}) = \prod_{r \in \mathcal{R}} Q(\mathbf{U}_{t}^{r})$.
This decouples all robots' planning, so that each robot updates and communicates their own control distribution $Q(\mathbf{U}_{t}^{r})$~(lines~\ref{alg:overview:brdcst} and~\ref{alg:overview:rcv}).
In line~\ref{alg:overview:CEM}, each robot runs an independent CEM loop that updates its own actions $\mathcal{Q}(\mathbf{U}^r_t)$ towards maximising the team's reward given communicated distributions of other robots.
The communication need not be synchronous, and line~\ref{alg:overview:rcv} returns only the latest distributions that are available. 

\begin{algorithm}[t!]
	\caption{Overview of decentralised receding-horizon planning for robot~$r$} \label{alg:overview}
    \textbf{Inputs:} Object measurements $\mathbf{y}_{t}^{B}$, reward function $R^r$ \\
    \textbf{Outputs:} Control for robot~$r$ $\mathbf{U}_{t}^{r}$, 
	\begin{algorithmic}[1]    
    \State $\mathcal{B}_{t}(\mathbf{O}) \gets \texttt{update\_belief}(\mathcal{B}_{t-1}(\mathbf{O}), \mathbf{x}_{t}^{B}, \mathbf{y}_{t}^{B})$ \label{alg:overview:belief}
    \State $\mathcal{Q}(\mathbf{U}^r_t) \gets \mathcal{N}(\mathbf{0}, \sigma_{0}^{2}I)$ \label{alg:cem:init}    
	\For{fixed number of iterations} \label{alg:overview:iter}
        \State $\mathcal{Q}(\mathbf{U}_t^{\mathcal{R}\setminus r}) \gets \texttt{receive\_distribution()}$ \label{alg:overview:rcv}
        \State $\mathcal{Q}(\mathbf{U}_t^r) \gets \texttt{dec\_cem}(R^r,\mathcal{Q}(\mathbf{U}_t^{\mathcal{R}\setminus r}))$ \label{alg:overview:CEM}
        \State $\texttt{broadcast\_distribution}(\mathcal{Q}(\mathbf{U}_t^r))$ \label{alg:overview:brdcst}
    \EndFor
    \Return $\mathbf{U}_t^{r} \sim \mathcal{Q}(\mathbf{U}_t^r)$
	\end{algorithmic} 
\end{algorithm}

\begin{algorithm}[t!]
	\caption{Procedure \texttt{dec\_cem} for robot $r$} \label{alg:cem}
    \textbf{Inputs:} Reward function $R^r$, other robots' control distributions $\mathcal{Q}(\mathbf{U}_t^{\mathcal{R}\setminus r})$\\
    \textbf{Outputs:} Control distribution for robot $r$, $\mathcal{Q}(\mathbf{U}_t^r)$
	\begin{algorithmic}[1]    
    \For{fixed number of iterations}
        \State $\mathbf{\hat{U}}^r_t \sim \mathcal{Q}(\mathbf{U}^r_t)$ {\tt{//Sample own controls}}\label{alg:cem:sample}  
        \State $\mathbf{\hat{U}}^{\mathcal{R}\setminus r}_t \sim \mathcal{Q}(\mathbf{U}_t^{\mathcal{R}\setminus r})$ {\tt{//Sample others'}}\label{alg:cem:sample_other}    
        \Statex {\tt{//Compute own elite set}}
        \State $\mathbf{\hat{U}}^{r*}_t = \{\mathbf{U}_t \in \mathbf{\hat{U}}^r_t \mid R^r(\mathbf{\hat{U}}^r_t, \mathbf{\hat{U}}^{\mathcal{R}\setminus r}_t) > \underbar{R}\}$ \label{alg:cem:elite}
        \Statex {\tt{//Update own distribution}}
        \State $\mathcal{Q}(\mathbf{U}^r_t) \gets \texttt{fit\_gaussian}(\mathbf{\hat{U}}^{r*}_t)$ \label{alg:cem:fit}
        \EndFor
\Return $\mathcal{Q}(\mathbf{U}^r_t)$
	\end{algorithmic} 
\end{algorithm}

\subsection{Belief Update Procedure}

Given measurements $\mathbf{Y}_t^{B}$, we use the information form of the Kalman filter to obtain a Gaussian belief over objects~$\mathcal{B}_t(\mathbf{O}) = \mathcal{N}(\hat{\mathbf{O}}_{t}, \Lambda_{t}^{-1})$ with mean $\hat{\mathbf{O}}_{t}$ and information matrix $\Lambda_{t}$, which is the inverse of covariance. 
The benefit of using the information form is that belief updates are \emph{additive}.
\begin{equation}\label{info_filter}
\begin{aligned}
    \Lambda_{t} &= \Lambda_{t-1} + \sum_{\beta \in B} \mathbf{I}(\mathbf{x}_{t}^{\beta}), \\ 
    \hat{\mathbf{O}}_{t} &= \Lambda_{t}^{-1}( \Lambda_{t-1} \hat{\mathbf{O}}_{t-1} + \sum_{\beta \in B} \mathbf{I}(\mathbf{x}_t^{\beta})\mathbf{y}_{t}^{\beta} ).
\end{aligned}
\end{equation}
Here, $\mathbf{I}(\mathbf{x}_{t}^{\beta})$ is the innovation matrix. 
We account for the sensing range constraint~\eqref{eq:sensor_model} by setting $\mathbf{I}(\mathbf{x}_{t}^{\beta}) = \sigma^{-2} I$ if $\mathbf{y}_{t}^{\beta}$ is within sensing range, and $0$ otherwise (i.e. the measurement is spurious). 
This automatically ignores objects outside of the sensing range, and corresponds to having infinite measurement error. 

\subsection{Decentralised Cross Entropy Method}
Given the updated belief, the DecCEM procedure updates the control distribution of each robot $\mathcal{Q}(\mathbf{U}^r_t)$ towards maximising the team's reward.
In doing so, it uses other robots' control distributions $\mathcal{Q}(\mathbf{U}^{\mathcal{R} \setminus r}_t)$ that are periodically communicated (lines~\ref{alg:overview:rcv},~\ref{alg:overview:brdcst}, Alg.~\ref{alg:overview}).
The receipt of other robots' distributions may be asynchronous, and the algorithm gracefully degrades with loss of communication. 

A DecCEM iteration for a robot $r$ comprises the following.
First, random samples $\mathbf{\hat{U}}_{t}^{r}$ are drawn from the current distribution of its own controls (line~\ref{alg:cem:sample}). 
Additionally, random samples $\mathbf{\hat{U}}_{t}^{\mathcal{R}\setminus r}$ are drawn from the other robots' control distributions that are communicated (line~\ref{alg:cem:sample_other}).
The samples are propagated through the robot dynamic model and used to evaluate the team's reward function.
Using the reward values, an `elite set' $\mathbf{\hat{U}}^{r*}_t$ of robot $r$'s own control samples is extracted whose reward exceeds a set threshold $\underbar{R}$  (line~\ref{alg:cem:elite}). 
Subsequently, the robot $r$'s control distribution is updated by fitting a Gaussian to the elite set (line~\ref{alg:cem:fit}).
This process is repeated for a fixed number of iterations before broadcasting.
The outer loop of updating and communicating the distributions is repeated to yield the final control distribution. 

In our implementation, we improve the CEM iteration by extracting the best $N_e$~samples rather than using a threshold. 
Further, we terminate the iteration if the average variance of the control actions falls below a set threshold to prevent over-fitting.

Whilst we do not provide convergence guarantees, the computational complexity of DecCEM in Alg.~\ref{alg:cem} is a primarily a function of the number of iterations and choice of control distribution, $\mathcal{Q}(\mathbf{U}_t^r)$. In this paper we select a Gaussian distribution due to the high availability of efficient expectation maximisation algorithms.
\subsection{Reward Functions}\label{sec:contrib:escorter}
The reward function $R^r$ used for DecCEM differs for PAs and EAs. 
For PAs, the reward is set as simply the marginal log probability of task satisfaction given the current belief over objects:
\begin{equation}\label{eq:pa_reward}
    R^{A}(\mathbf{U}^{A}_{t}) = \log P(\phi \mid \mathbf{X}_t^A) = \log \mathbb{E}_{\mathbf{O} \sim \mathcal{B}_{t}(\cdot)}[P(\phi \mid \mathbf{X}_t^A, \mathbf{O})].
\end{equation}
Notably, the PA's reward function does not depend on future measurements gathered by the EA, which prompts the PAs to plan conservatively, given only the current belief.

Given the PAs' control distribution~$\mathcal{Q}(\mathbf{U}_{t}^{A})$ computed to maximise~\eqref{eq:pa_reward}, an EA should choose its own controls to best support the PAs in satisfying tasks. 
The EA's plan should therefore focus not solely on improving $\mathcal{B}$ over the entire environment $\mathcal{E}$ as is typical in information gathering \cite{atanasov2014, atanasov2015, schlotfeldt2018}, but rather on improving $\mathcal{B}$ to increase the PAs' probability of task satisfaction. 
Here, we present two reward functions that capture this logic, and one more traditional information gathering reward based on MI-UCB. 

\subsubsection{Satisfaction Improvement~(SI)}\label{sec:contrib:escorter:si}
The most immediate approach is for EAs to directly maximise the PAs' probability of task satisfaction.
To this end, the SI approach aims to maximally improve the PAs' expected probability of task satisfaction, through conditioning with EA measurements.
In other words, it solves:
\begin{equation}\label{eq:si_objective}
    R^{B}_{\textsc{SI}}(\mathbf{U}^{B}_{t}) = \mathbb{E}_{\mathbf{X}_{t}^{A} \sim \mathcal{Q}(\cdot) } \left[ P( \phi \mid \mathbf{X}_{t}^{A},  \mathbf{Y}_{t}^{B} )  - P( \phi \mid \mathbf{X}_{t}^{A} ) \right],
\end{equation}
where $\mathbf{X}_{t}^{A}$ is sampled implicitly from $\mathcal{Q}(\mathbf{U}_{t}^{A})$ through the dynamic model.

A challenging aspect of computing the SI objective~\eqref{eq:si_objective} is the implicit dependence between $\mathbf{U}_{t}^{B}$ and the conditioning measurements $\mathbf{Y}_{t}^{B}$. 
To alleviate this difficulty, we use the following rearrangement of the first term, which can be derived from Bayes' rule and the conditional independence properties in the PGM~(Fig.~\ref{fig:prob_form_diagram}):
\begin{equation}\label{eq:posterior_trick}
    P \left( \phi \mid \mathbf{X}_{t}^{A},  \mathbf{Y}_{t}^{B} \right) = \mathbb{E}_{\mathbf{O}\sim\mathcal{B}_{t}(\cdot\mid\mathbf{U}^{B}_t)} \left[ P(\phi\mid\mathbf{X}_t^{A}, \mathbf{O}) \right]
    .
\end{equation}
Here, $\mathcal{B}_{t}(\mathbf{O} \mid \mathbf{U}^{B}_t)$ is the \emph{predicted} belief over objects after the EA executes action $\mathbf{U}^{B}_t$. 
For the Gaussian targets we consider, such prediction can be achieved by propagating the information matrix~\eqref{info_filter} forward in time, while retaining the mean.
In other words, $\Lambda_{t+T} = \Lambda_{t} + \sum_{\tau\in T, \beta\in B} I(\mathbf{x}_{\tau}^{\beta})$. 

\subsubsection{Reduction in Satisfaction Entropy~(SE)}\label{sec:contrib:escorter:se}
Whereas the SI approach aims to simply increase the probability of satisfaction implied by the PAs' control distribution, one may argue that from an information gathering perspective, decreasing probability of task satisfaction is equally as valuable as increasing it. In other words, it may be of equal value to measure the change in task satisfaction.
To this end, we consider reducing the binary entropy of probability of satisfaction with the measurement set $\mathbf{Y}_{t}^{B}$.
Formally, the SE approach solves:
\begin{equation}\label{eq:task_entropy}
\text{
     \small{$R^{B}_{\textsc{SE}}(\mathbf{U}^{B}_{t}) =\mathbb{E}_{\mathbf{X}_{t}^{A} \sim \mathcal{Q}(\cdot) } \left[ h( P(\phi\mid\mathbf{X}_{t}^{A}) ) - h( P(\phi\mid\mathbf{X}_{t}^{A}, \mathbf{Y}_{t}^{B}) ) \right]$},
     }
\end{equation}
where $h(P) = -P \log P - (1 - P) \log P$ is the binary entropy.
The posterior probability of task satisfaction $P(\phi\mid\mathbf{X}_{t}^{\alpha})$ is calculated analogously to SI using~\eqref{eq:posterior_trick}. Here the reward favours lower entropy after EA measurements. Since binary entropy takes its maximum of 1 at $P = 1/2$ and is otherwise symmetric decreasing around this point, the SE objective either increases or decreases the probability of satisfaction of the samples drawn from the EAs, towards greater certainty. 

\begin{figure}[t]
    \centering
    \includegraphics[width=0.95\columnwidth]{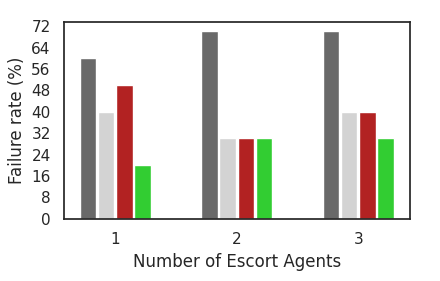}
    \vspace{-2ex}
    \caption{PA's failure rate with different EA reward variants. Higher is worse. Failure is recorded when the PA collides with an object. \textbf{Dark grey:} blind, \textbf{light grey:} MI-UCB, \textbf{red:} SI, \textbf{green:} SE.}
    \vspace{-3ex}
    \label{fig:percentage_of_collision}
\end{figure}

\begin{figure*}[!htbp]
    \centering
    \vspace{2ex}
    \setlength{\tabcolsep}{0em} 
  \begin{adjustbox}{valign=t}
    \begin{tabular}[!htbp]{ccccc}
    \rotatebox[origin=c]{90}{\textbf{Blind}} &
    \raisebox{-0.5\height}{\includegraphics[width=0.2301\linewidth]{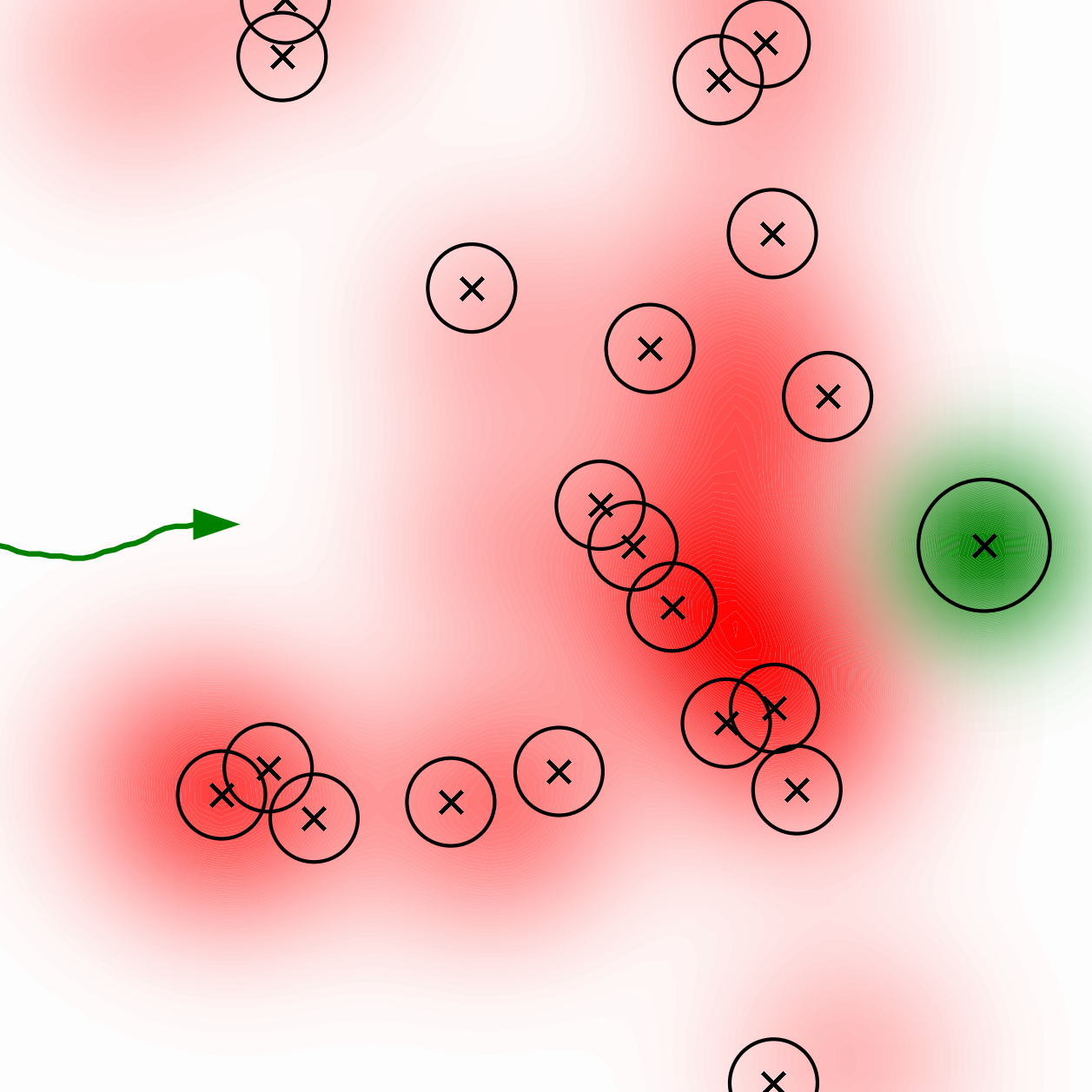}} &
    \raisebox{-0.5\height}{\includegraphics[width=0.2301\linewidth]{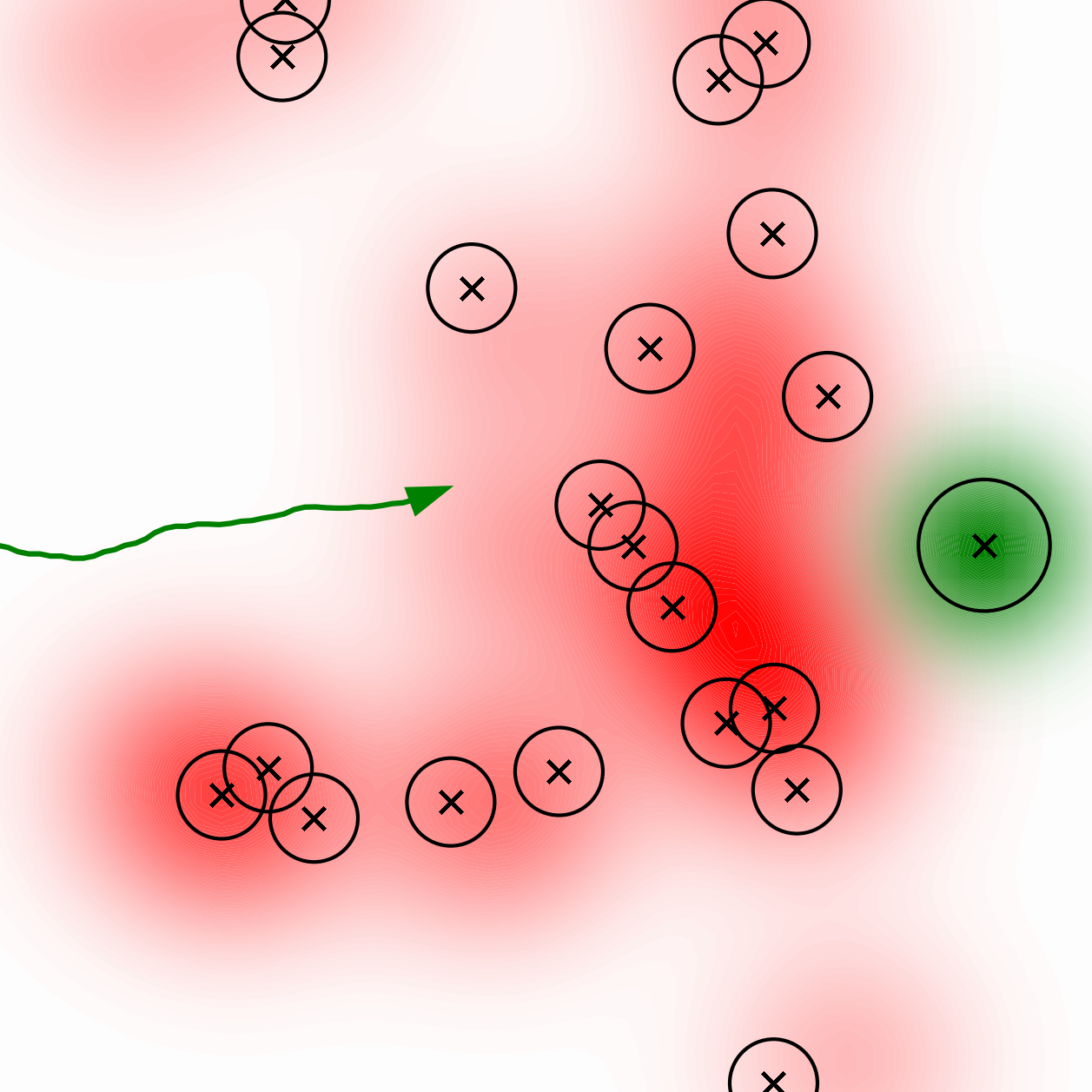}} &
    \raisebox{-0.5\height}{\includegraphics[width=0.2301\linewidth]{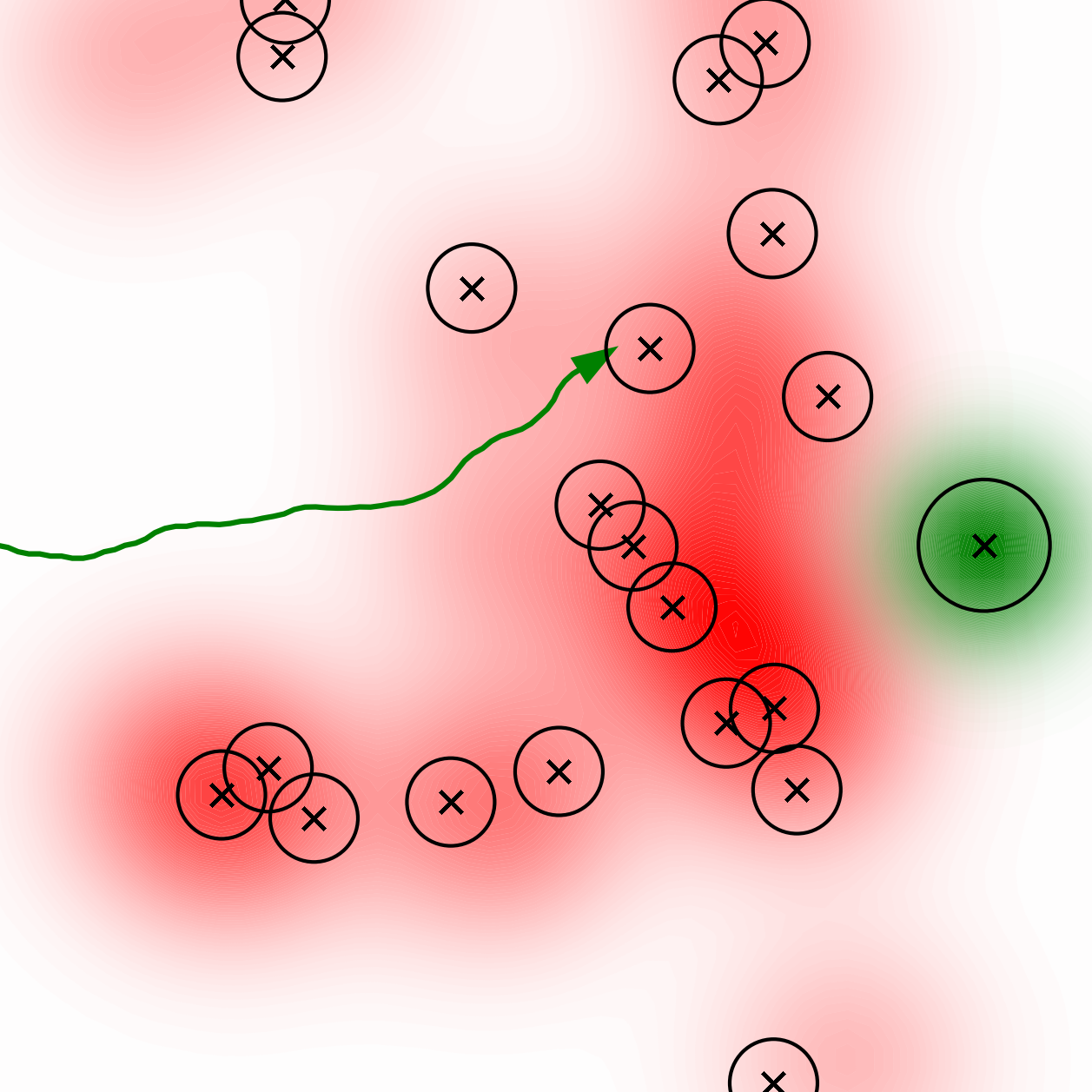}} &
    \raisebox{-0.5\height}{\includegraphics[width=0.2301\linewidth]{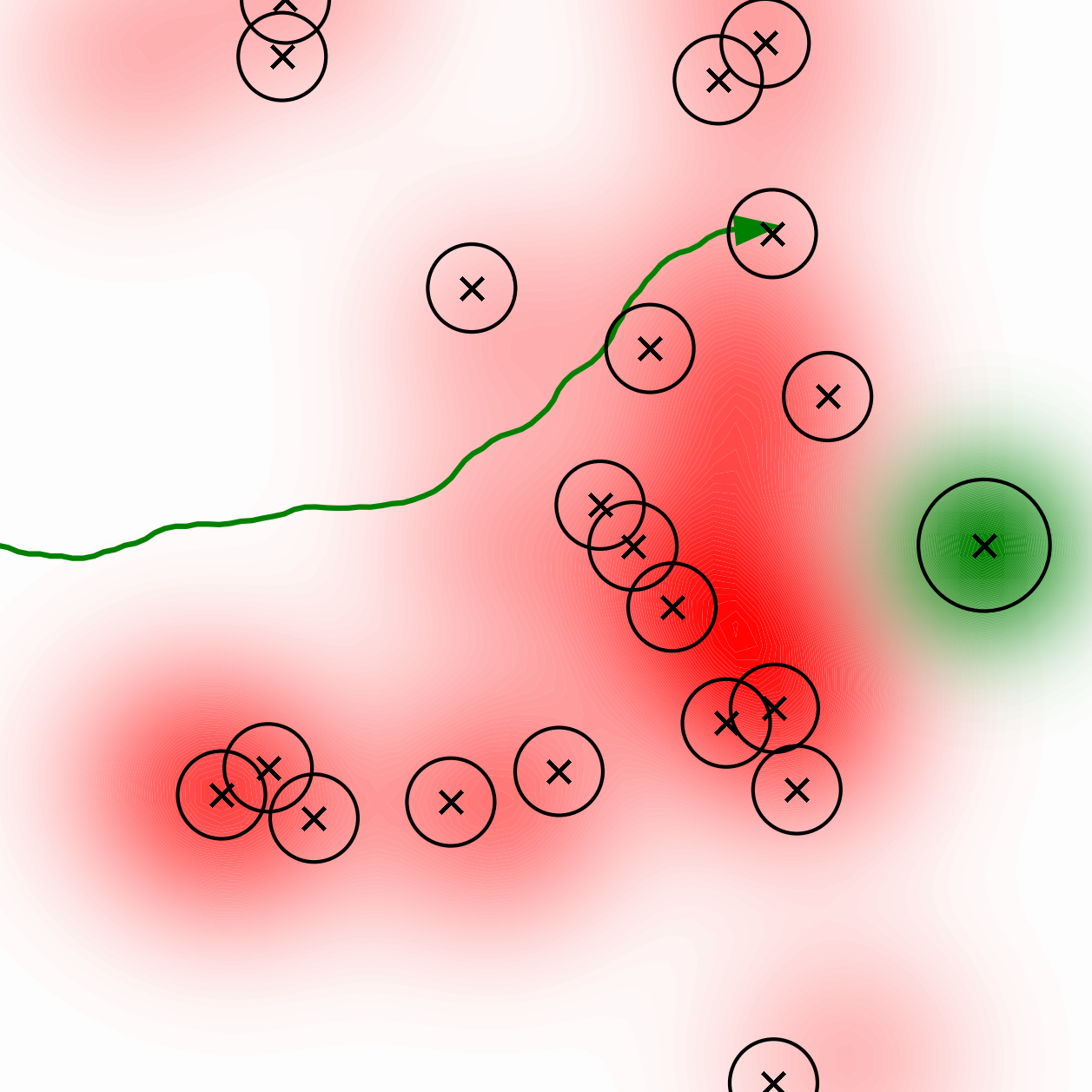}} \\ 
    \rotatebox[origin=c]{90}{\textbf{MI-UCB}} &
    \raisebox{-0.5\height}{\includegraphics[width=0.2301\linewidth]{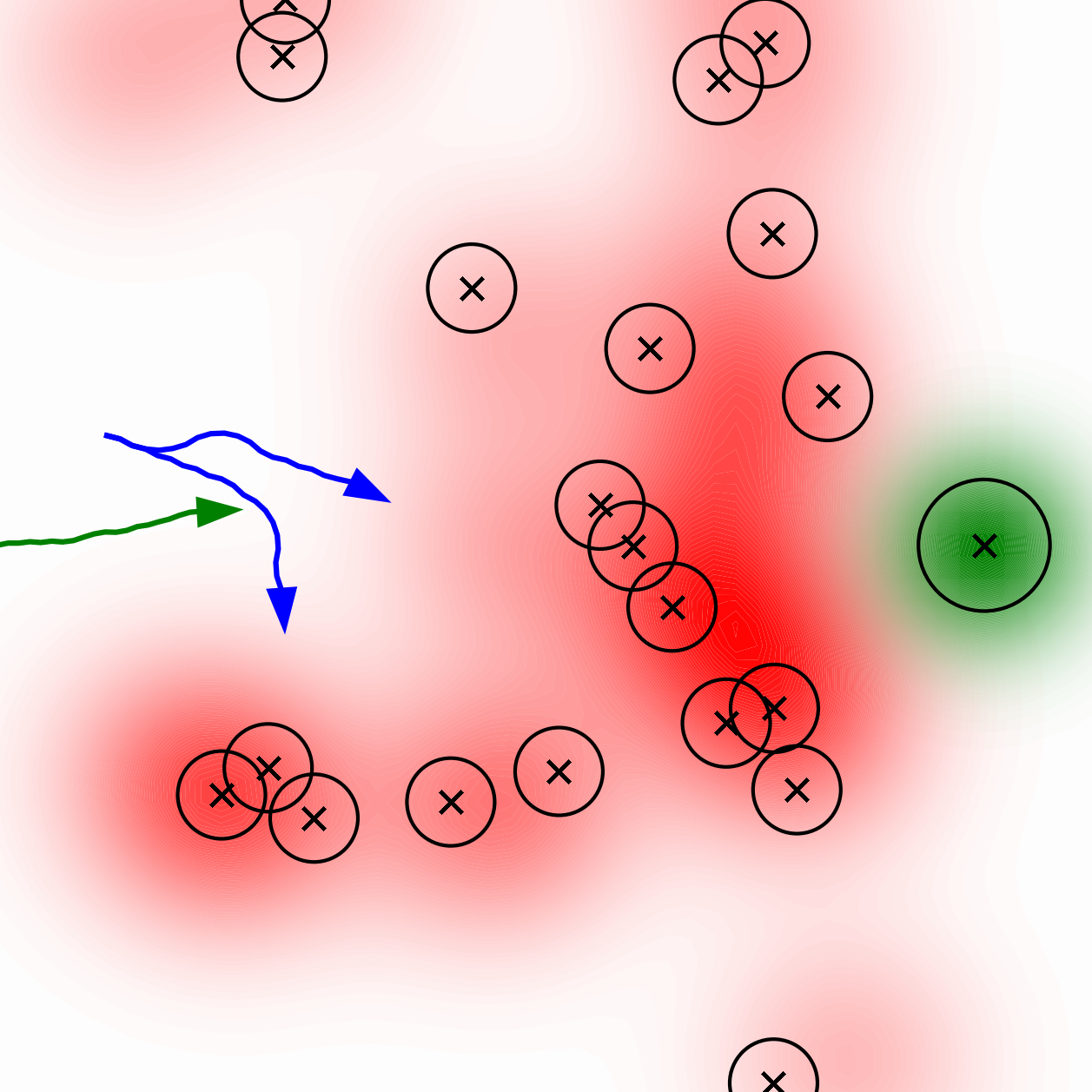}} &
    \raisebox{-0.5\height}{\includegraphics[width=0.2301\linewidth]{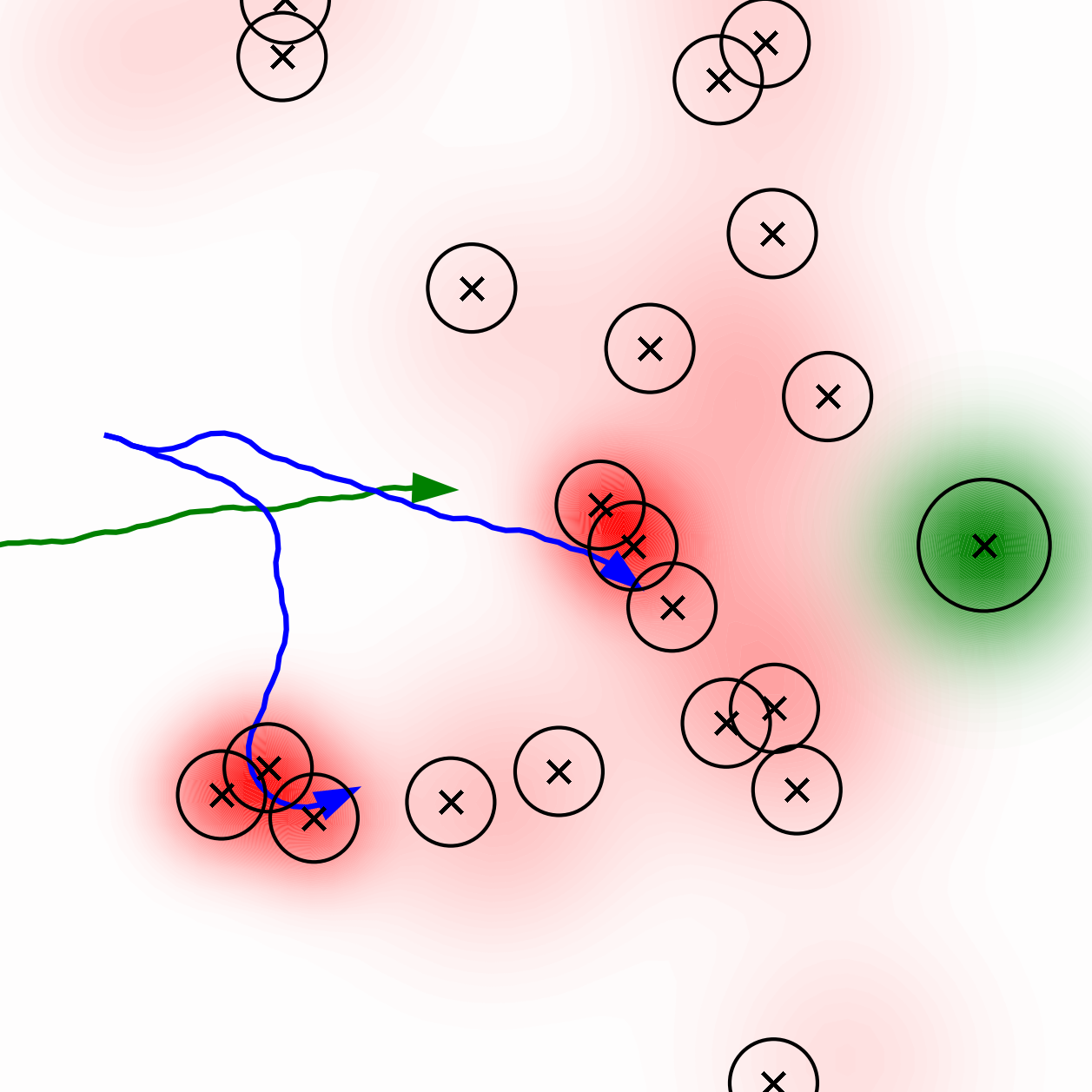}} &
    \raisebox{-0.5\height}{\includegraphics[width=0.2301\linewidth]{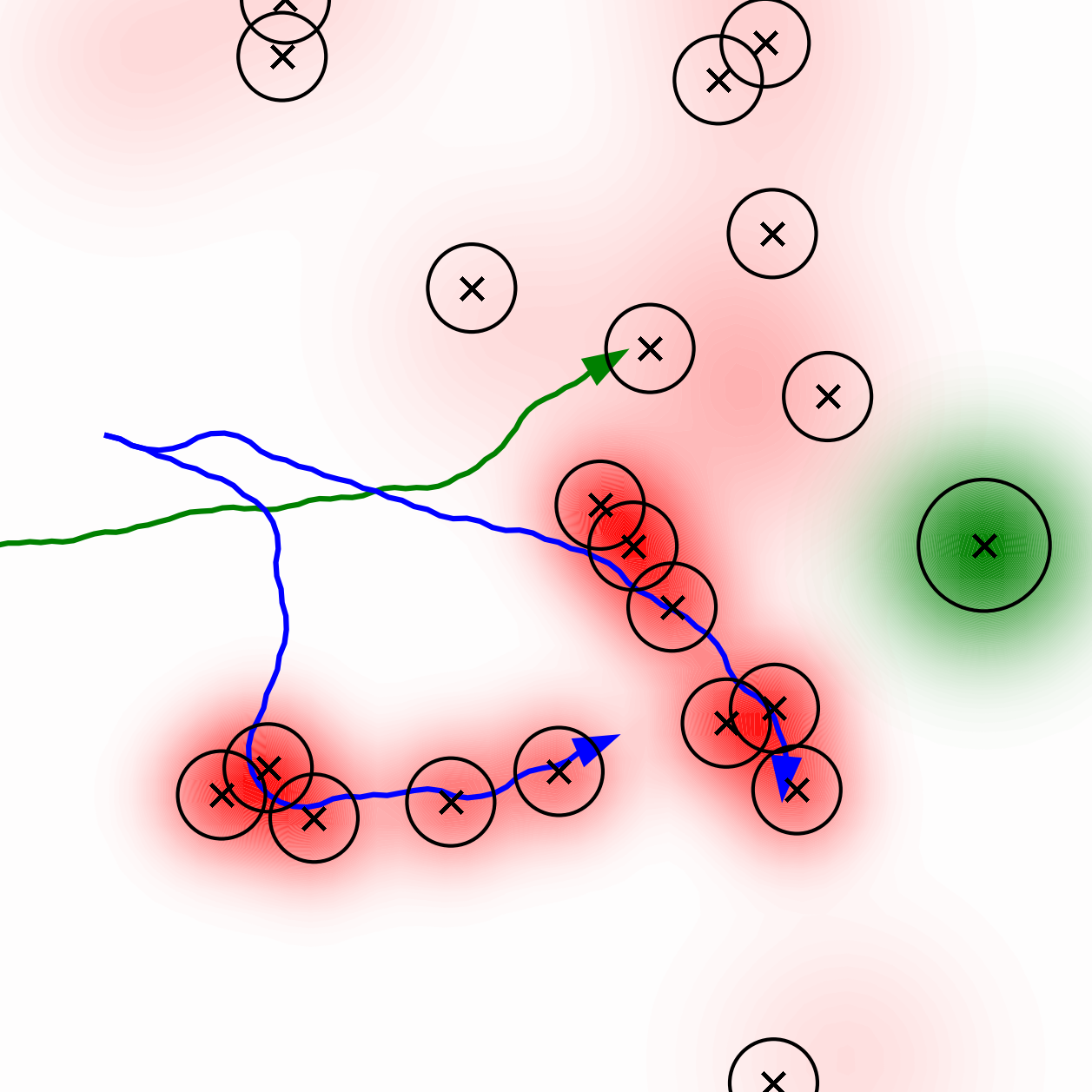}} &
    \raisebox{-0.5\height}{\includegraphics[width=0.2301\linewidth]{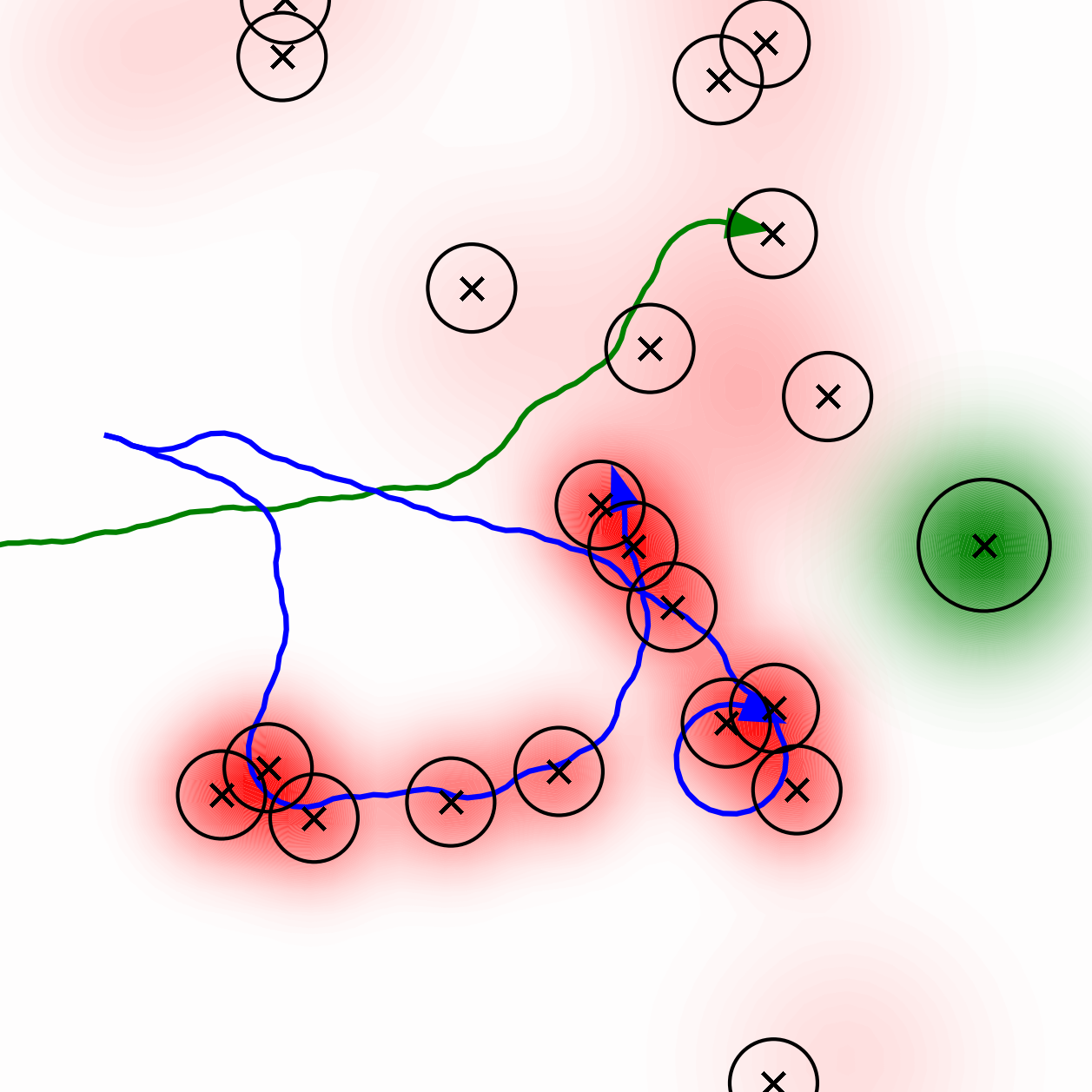}} \\
    \rotatebox[origin=c]{90}{\textbf{SE}} &
    \raisebox{-0.5\height}{\includegraphics[width=0.2301\linewidth]{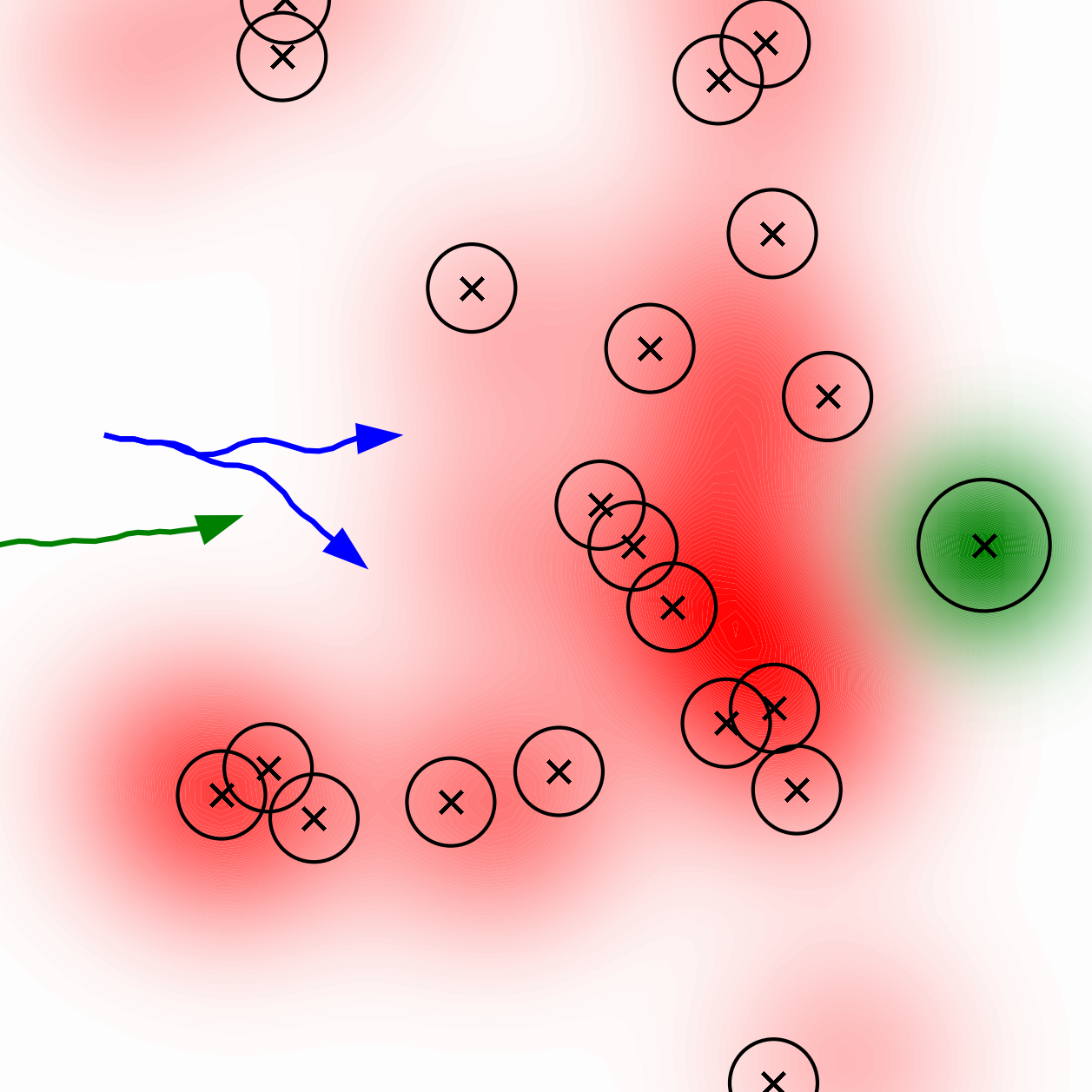}} &
    \raisebox{-0.5\height}{\includegraphics[width=0.2301\linewidth]{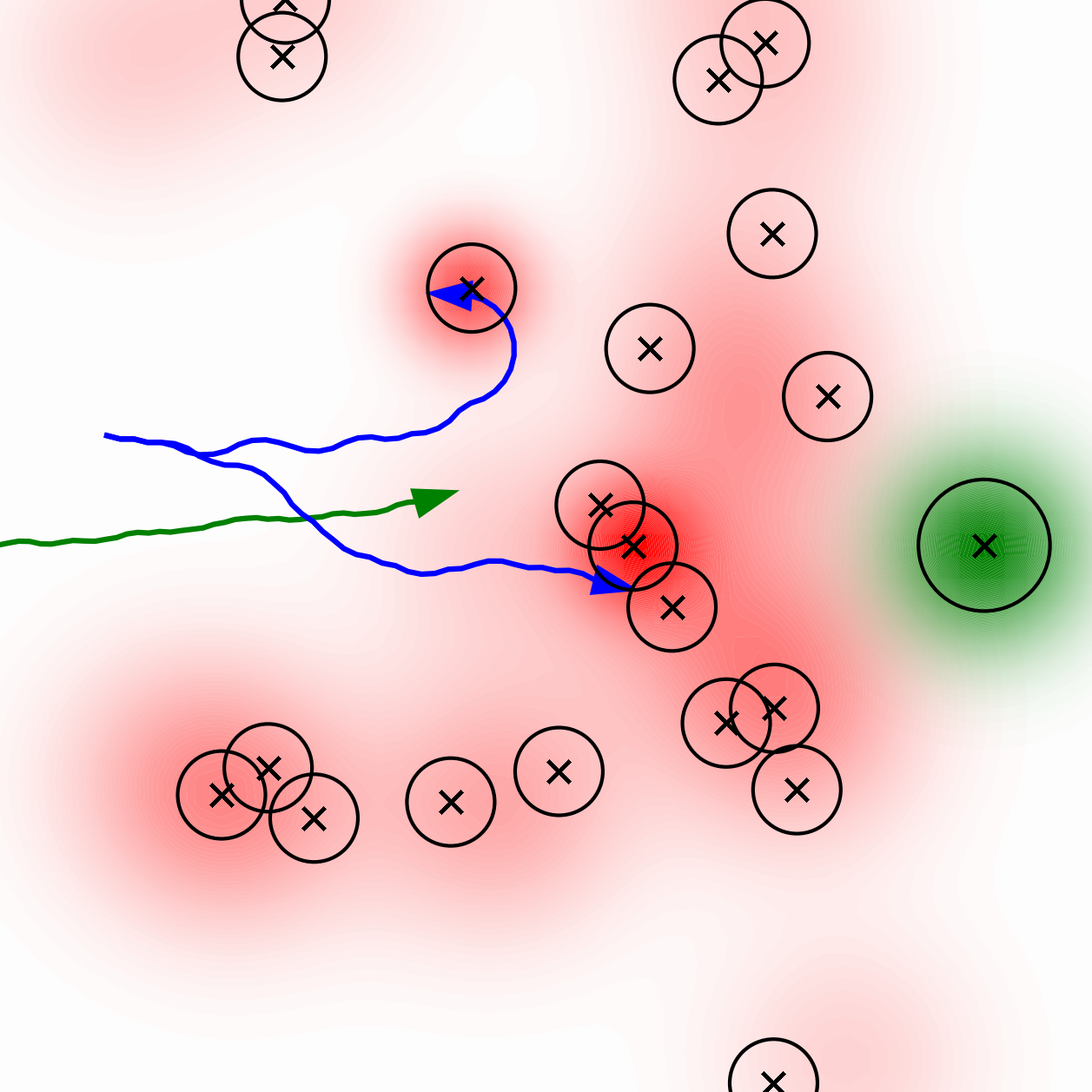}} &
    \raisebox{-0.5\height}{\includegraphics[width=0.2301\linewidth]{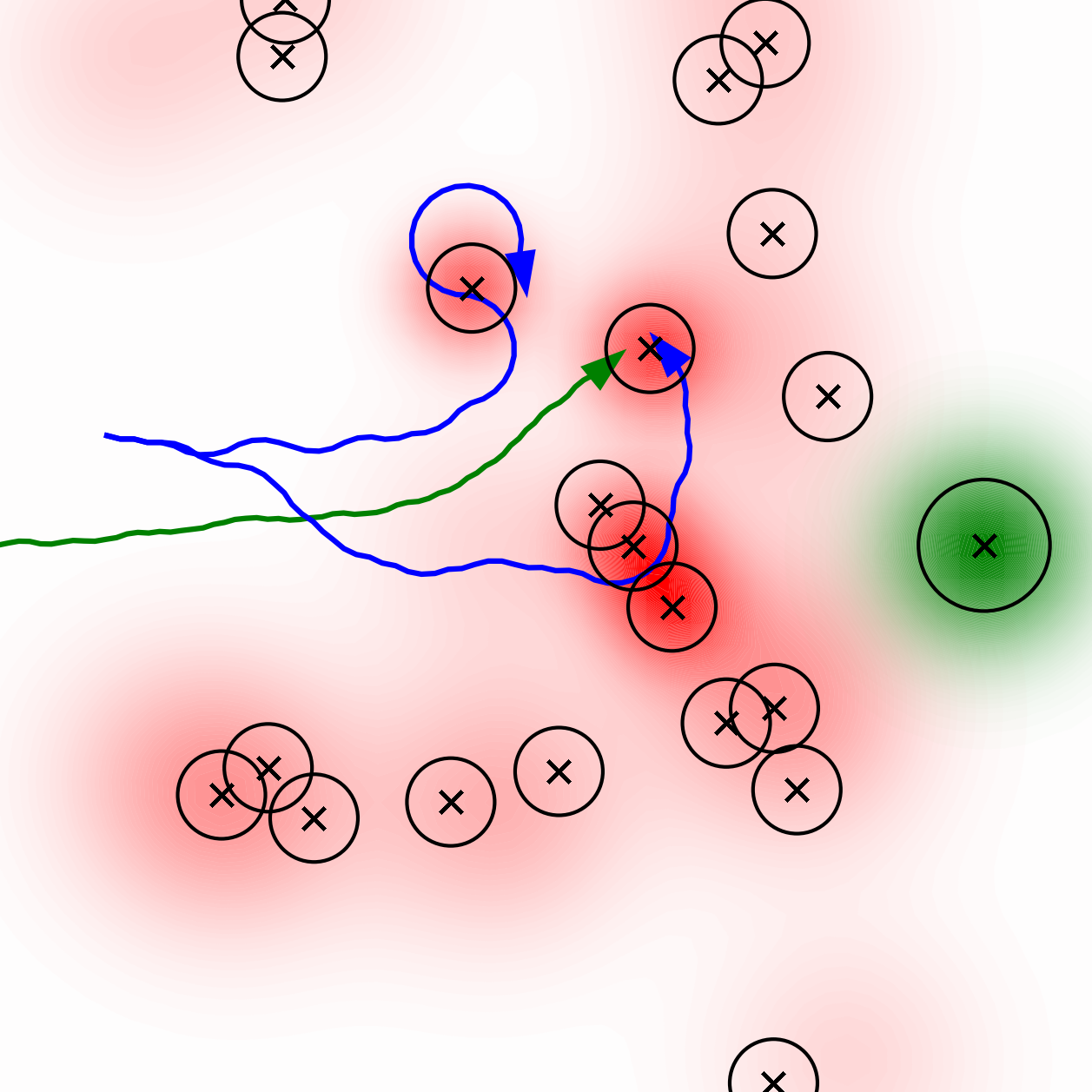}} &
    \raisebox{-0.5\height}{\includegraphics[width=0.2301\linewidth]{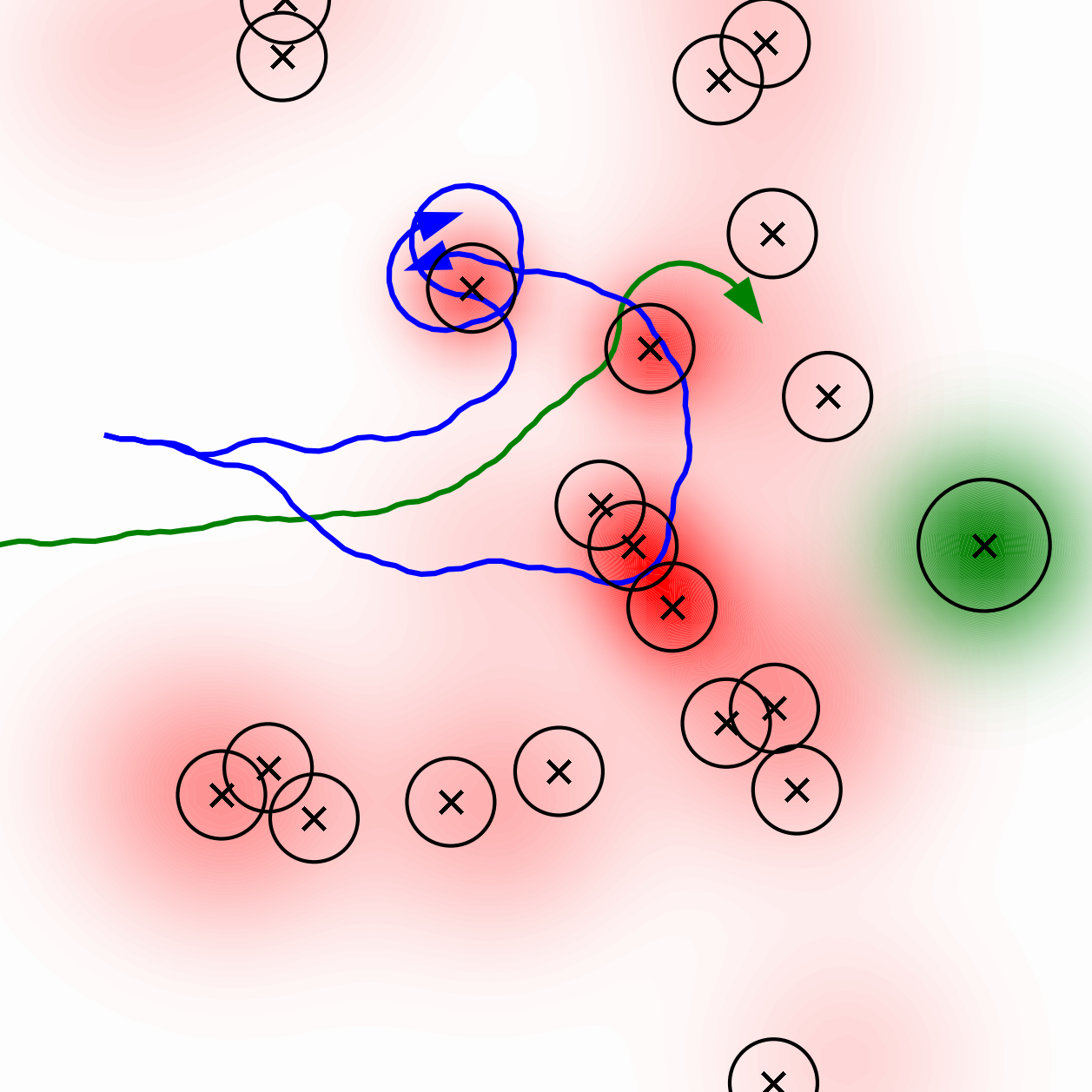}} \\
    \rotatebox[origin=c]{90}{\textbf{SI}} &
    \raisebox{-0.5\height}{\includegraphics[width=0.2301\linewidth]{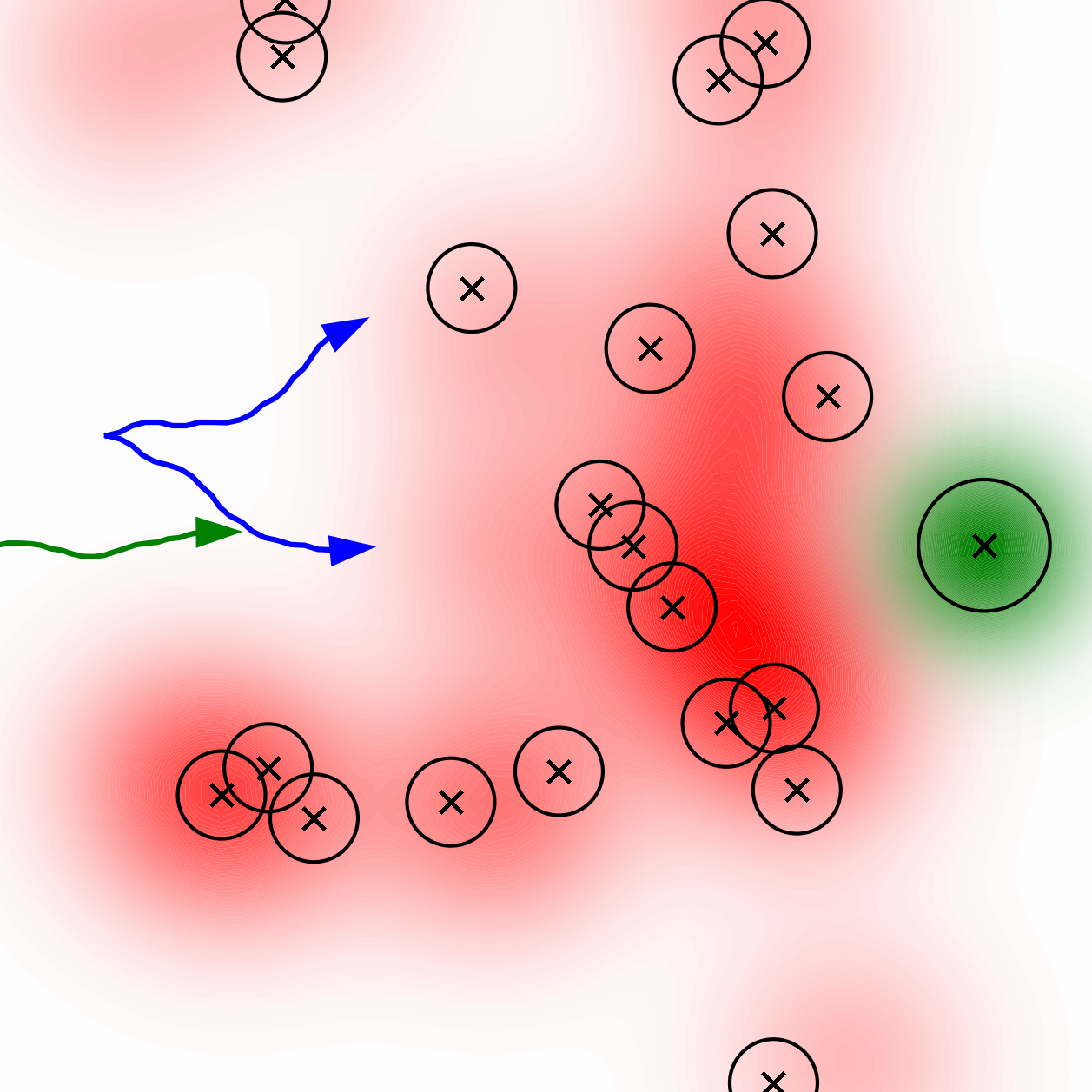}} &
    \raisebox{-0.5\height}{\includegraphics[width=0.2301\linewidth]{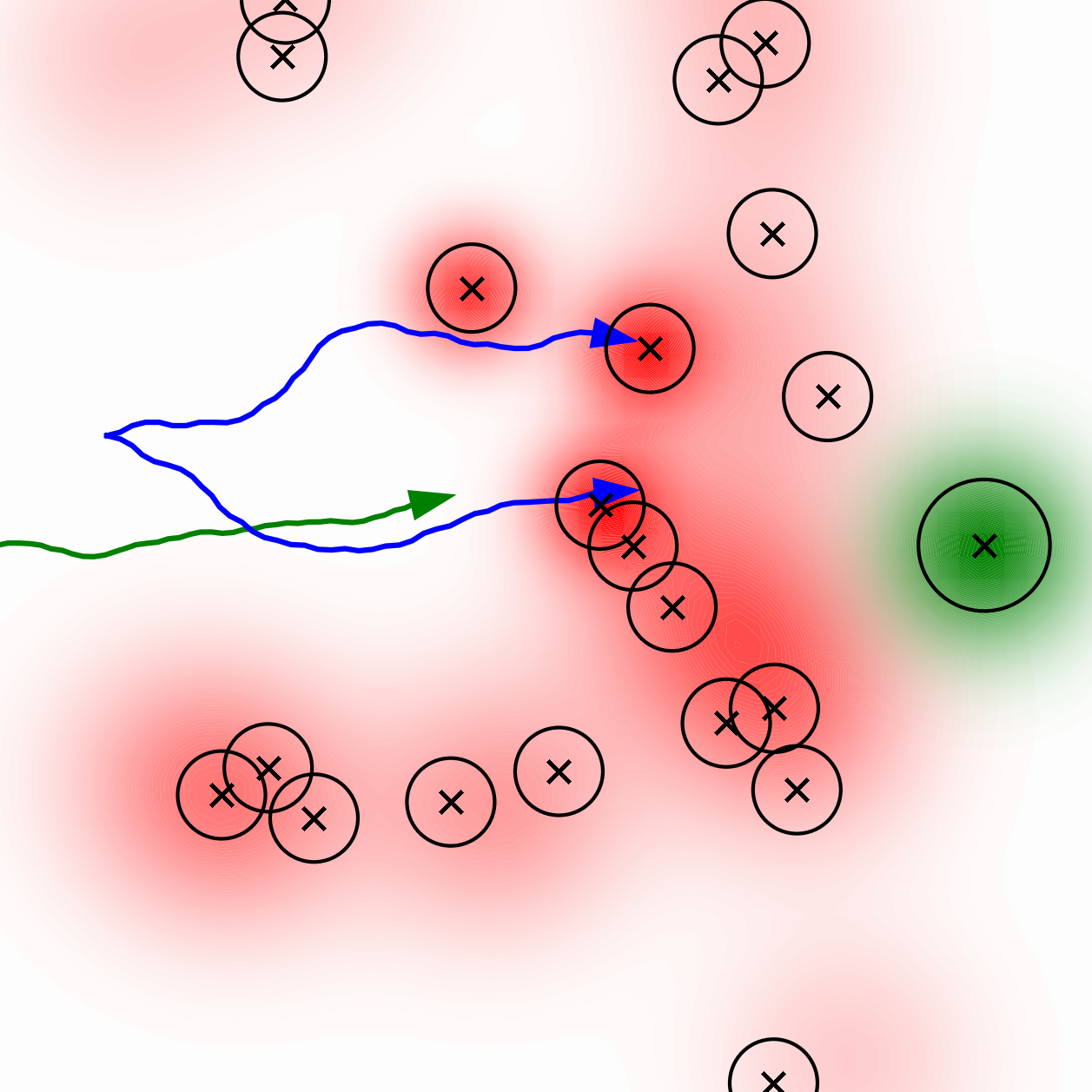}} &
    \raisebox{-0.5\height}{\includegraphics[width=0.2301\linewidth]{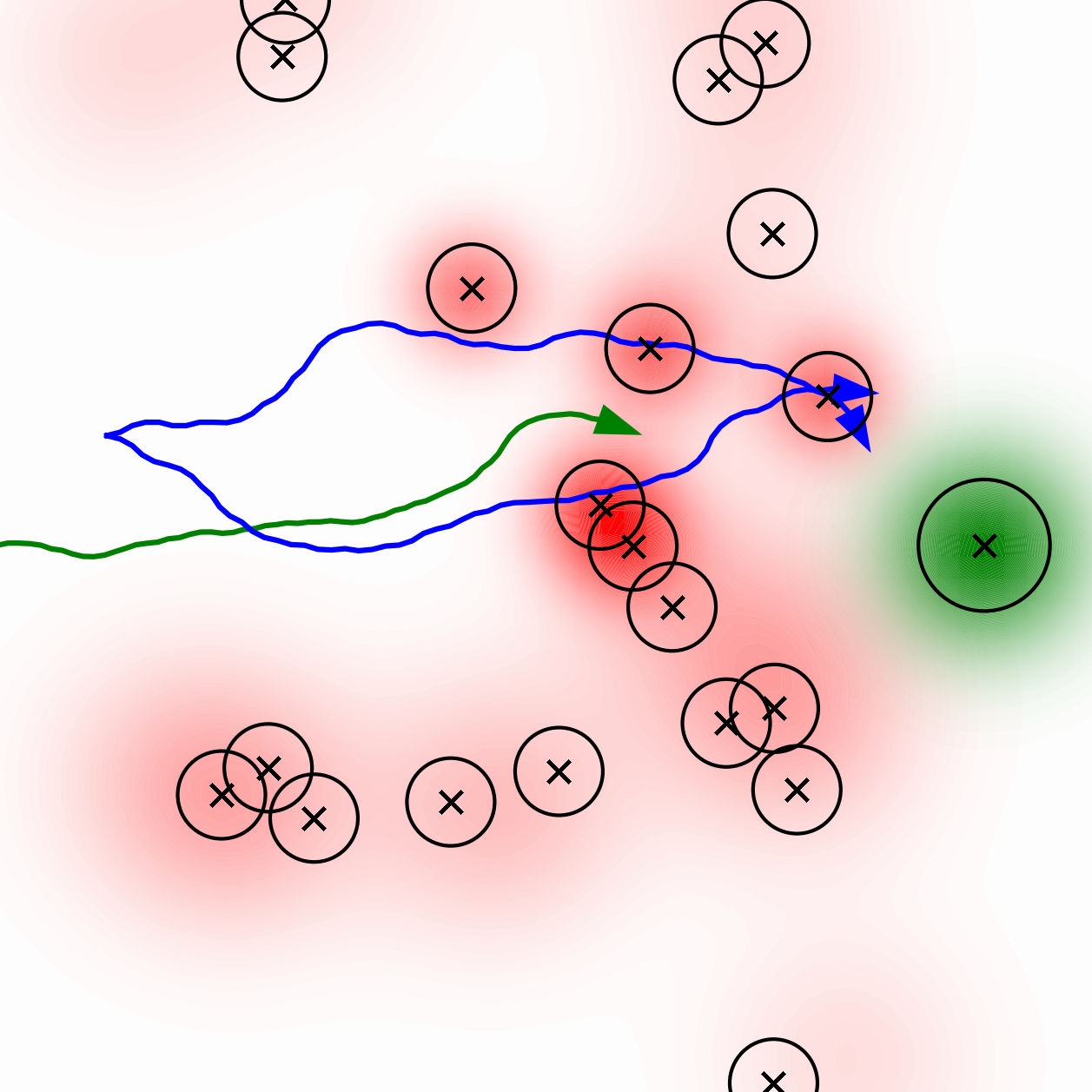}} &
    \raisebox{-0.5\height}{\includegraphics[width=0.2301\linewidth]{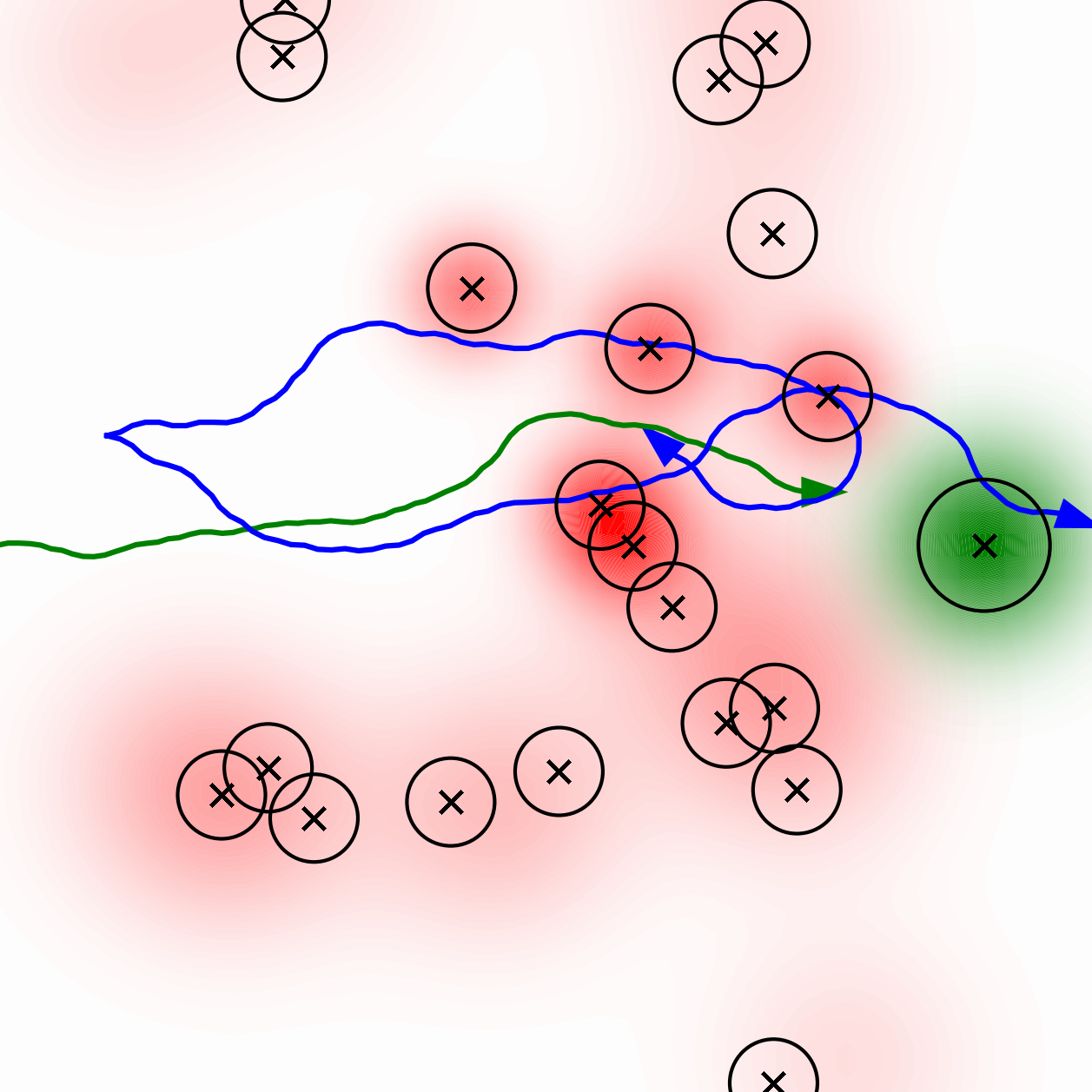}} \\
    & $t = 10$ & $t = 20s$ & $t = 30s$ & $t = 40s$
    \end{tabular}
    \end{adjustbox}
    \caption{Comparison of the 2D object avoidance for the four reward functions with two escort agents (blue) and one principal agent (green). The solid lines represent the path history and the triangle represents the robots position and heading at the given time $t$. The ground truth position of the objects are represented with an `x', and the circle represents the edge of those objects. The destination, $d$, is represented by the green heat-map. The red heat-map represents the belief of those obstacles. \textbf{Note}: the colour intensity is normalised to the maximum probability of belief (i.e., observing the obstacle does not affect the belief probability in other areas, despite appearing decreased).}
    \label{fig:4square_cmp}
\end{figure*}

\subsubsection{MI-UCB}\label{sec:contrib:escorter:mi_ucb}
The MI-UCB provides an upper bound on the posterior task satisfaction in terms of prior expected task satisfaction and information gain. 
Using MI-UCB, it can be shown that, with probability~$\geq 1 - \delta$~\cite{brian2021}:
\begin{equation} \begin{aligned}\label{eq:mi_ucb}
    P(\phi\mid&\mathbf{X}_{t}^{A}, \mathbf{Y}_{t}^{B})   \\
    &\leq \log \mathbb{E}_{\mathbf{O} \sim \mathcal{B}_{t}(\cdot)} \exp P(\phi\mid\mathbf{X}_{t}^{A} ,\mathbf{O}) + \frac{1}{\delta}I(\mathbf{Y}_{t}^{B}; \mathbf{O})
    .
\end{aligned} \end{equation}
Here, the expectation in the first term is taken with respect to the current belief, and $I(\mathbf{Y}_{t}^{B}; \mathbf{O})$ is the Shannon information gain between the measurements $\mathbf{Y}_{t}^{B}$ and object locations $\mathbf{O}$. 
For Gaussian targets considered in this paper, the information gain is given by $I(\mathbf{Y}_{t}^{B}; \mathbf{O}) = \frac{1}{2} ( \log \det \Lambda_{t+T} - \log \det \Lambda_{t})$ in terms of information matrices.

A striking feature of the MI-UCB~\eqref{eq:mi_ucb} is that the PAs and EAs are fully decoupled, in the sense that their rewards simply add up.
Since the EAs' actions only affect the information gain term, it is the only reward for the EAs to maximise. 
Hence the MI-UCB approach solves:
\begin{equation}
    R^{B}_{\textsc{MI-UCB}}(\mathbf{U}^{B}_{t}) = I \left( \mathbf{Y}_{t}^{B}; \mathbf{O} \right)
    .
\end{equation}
That is, the MI-UCB objective is equivalent to information maximisation commonly considered in the literature~\cite{atanasov2014,atanasov2015}.

\section{Results}


\subsection{Simulation setup}
In the following simulated results, the PAs plan toward achieving a task $\phi$ with probability of success $P(\phi \mid \mathbf{X}_{t}^{\alpha}, \mathbf{O})$ given PAs' trajectory and objects $\mathbf{O}$. 
We consider a subset of task classes where the objective is to reach a destination~$\mathbf{d}$ while avoiding objects~$\mathbf{O}$.
Inspired by~\cite{rstl_brian2021}, we model this reach-avoid task as a conjunction of reach and avoid tasks:
\begin{equation}\label{eq:psat}
    P( \phi \mid \mathbf{X}_{t}^{\alpha}, \mathbf{O} ) = \prod_{\tau, i} P \left( \phi^{O} \mid \mathbf{x}_{\tau}^{\alpha}, \mathbf{o}_{i} \right) P \left( \phi^{D} \mid \mathbf{x}_{\tau}^{\alpha} \right)
    ,
\end{equation}
with the reach ($\phi^{D}$) and avoid ($\phi^{O}$) tasks modelled as:
\begin{equation}
\begin{aligned}
    P( \phi^{O} \mid \mathbf{x}_{t}^{\alpha},  \mathbf{o}_{i} ) &= 1 - P_{O} \exp \left( - \frac{|| \mathbf{x}_{t}^{\alpha} - \mathbf{o}_{i} ||^{2}} {2 r_{O}^{2}} \right), \\
    P( \phi^{D} \mid \mathbf{x}_{t}^{\alpha})  &\propto \exp \left(- \frac{||  \mathbf{x}_{t}^{\alpha}- \mathbf{d} || ^{2}}{2 r_{D}^{2} } \right),
\end{aligned}
\end{equation}
and~$r_{D,O}$ are parameters that control the acceptance/collision radius, and $P_{O}$ controls the peak probability of collision.

All agents follow the bicycle kinematic model:
\begin{equation}
    \label{eq:kinematic_model}
    \begin{bmatrix}
        \dot{x} \\
        \dot{y} \\
        \dot{\theta}
    \end{bmatrix} = 
    \begin{bmatrix}
        v \cos\theta \\
        v \sin\theta \\
        u
    \end{bmatrix},
\end{equation}
where $x, y$ and $\theta$ are the position and heading respectively. 
$v$ and  $u$ denote the linear and angular speed.
$v$ is fixed separately for each agent, and $u$ is controllable.

Simulations are performed in environments of size $100 \times 100 m$, each containing 20 objects spawned uniformly randomly in the box $[20, 80] \times [20, 80]$.
All robots begin at $[10, 50]$. The PA's goal location is $[90, 50]$. 
The belief over each target is initialised with a high variance of $25 m^{2}$, and the initial mean is corrupted correspondingly.
The PA travels at $v_{PA} = 2 m s^{-1}$, and the EAs at $v_{EA} = 4 m s^{-1}$.

\subsection{Demonstration and Comparison}

We validate our approach in Sec.~\ref{sec:contrib:escorter} by examining PA task success rate over repeated simulations with each EA reward.
A `blind' variant consisting solely of the PA is introduced as a benchmark.
The blind PA plans a path given only the prior belief, without any measurements from the EAs.
For fair comparison, we generated 10 random environments and examined the percentage of failed PA trajectories, where failure is defined as trajectories that collided with objects in the ground truth. The result is shown in Fig.~\ref{fig:percentage_of_collision}.

Figure~\ref{fig:percentage_of_collision} shows that the SE approach~(Sec.~\ref{sec:contrib:escorter:si}) generally performs best, followed by MI-UCB and SI~(Sec.~\ref{sec:contrib:escorter:mi_ucb} and~\ref{sec:contrib:escorter:se}), and then blind.
Unsurprisingly, the blind approach performs the worst with the highest failure rate across all configurations due to the lack of resolution of environmental uncertainty.
The SE approach, in particular, consistently offers more than two-fold improvement over the blind benchmark.
Compared against the MI-UCB approach, SE performs better for lower number of escorts, and equally for higher number of robots.
This is because the SE approach makes better use of limited information by guiding the EAs towards objects that are more relevant to task.
The explicit consideration of PA's task performance leads to tighter coordination, and hence better performance, than the exploration-exploitation balance achieved by MI-UCB. 
Interestingly, the SI approach performs equally or worse than the MI-UCB approach.
This is because the SI approach only tries to \emph{validate} the current control distribution of the PAs, and hence cannot distinguish plans that, in fact, fail. 

To better understand the trends observed, we examine the system behaviour over time, illustrated in Fig.~\ref{fig:4square_cmp}.
The red heat-map represents the belief over obstacles (black crosses), normalised to the highest probability of belief. 
Consequently, when obstacles are observed the belief on the rest of the environment appears decreased whilst they stay the same.

The poor performance of the blind PA in Fig.~\ref{fig:percentage_of_collision} is exemplified by the failure at $t=30s$ in the first row.
The failure occurs because the blind PA is blocked by the large initial uncertainty of the objects and corrupted mean, which remain unresolved, and must therefore opt for a path that is not sufficiently object-free.
In other words, uncertainty resolution is necessary for better planning performance.

With the MI-UCB approach (second row), the EAs act towards reducing the uncertainty of targets. 
In doing so, at $t = 30s$, the EAs cover a cluster of obstacles that are close to the PA's trajectory as they are information rich, however fail to consider the information specific to the current PA trajectory, resulting in a similar collision to the blind case.

With the SE and SI approaches, the EAs generally reduce the uncertainty of targets that are closer to the PA's trajectory. 
In particular, it can be seen that the SI approach covers nearby obstacles at $t=30s$ allowing the PA to avoid them just beyond the collision range.

\section{CONCLUSION AND FUTURE WORK}
We presented the coordinated escort problem, a novel formulation of a joint optimisation problem where the coordination objective is dependent on the actions of an independent agent. We proposed DecCEM, a novel, decentralised solution to the joint, continuous control problem of coordinated escort. Based on the product-distribution approximation, DecCEM admits arbitrary parameterisation of the total control distribution, thus extending existing product-distribution based planners from discrete to continuous control spaces. We proposed and evaluated two new information gathering objective functions for escort agents whose mission is to increase probability of task satisfaction for PAs. These objective functions exhibited improved or comparable task success compared to general information gathering objectives. Future work will address the coordinated escort problem with dynamic adversarial objects, theoretical analysis of the convergence and optimality of DecCEM.



\bibliography{references.bib}

\begin{thebibliography}{10}
\providecommand{\url}[1]{#1}
\csname url@samestyle\endcsname
\providecommand{\newblock}{\relax}
\providecommand{\bibinfo}[2]{#2}
\providecommand{\BIBentrySTDinterwordspacing}{\spaceskip=0pt\relax}
\providecommand{\BIBentryALTinterwordstretchfactor}{4}
\providecommand{\BIBentryALTinterwordspacing}{\spaceskip=\fontdimen2\font plus
\BIBentryALTinterwordstretchfactor\fontdimen3\font minus
  \fontdimen4\font\relax}
\providecommand{\BIBforeignlanguage}[2]{{%
\expandafter\ifx\csname l@#1\endcsname\relax
\typeout{** WARNING: IEEEtran.bst: No hyphenation pattern has been}%
\typeout{** loaded for the language `#1'. Using the pattern for}%
\typeout{** the default language instead.}%
\else
\language=\csname l@#1\endcsname
\fi
#2}}
\providecommand{\BIBdecl}{\relax}
\BIBdecl

\bibitem{cem_kroese2006}
D.~P. Kroese, S.~Porotsky, and R.~Y. Rubinstein, ``The cross-entropy method for
  continuous multi-extremal optimization,'' \emph{Methodol. Comput. Appl.
  Probab.}, vol.~8, no.~3, pp. 383--407, 2006.

\bibitem{cem_for_planning}
M.~Kobilarov, ``Cross-entropy motion planning,'' \emph{Int. J. of Rob. Res.},
  vol.~31, no.~7, pp. 855--871, 2012.

\bibitem{best2019decmcts}
G.~Best, O.~Cliff, T.~Patten, R.~R. Mettu, and R.~Fitch, ``{Dec-MCTS}:
  Decentralized planning for multi-robot active perception,'' \emph{Int. J.
  Robot. Res.}, vol.~38, no. 2-3, pp. 316--337, 2019.

\bibitem{wolpert}
D.~H. Wolpert and C.~E.~M. Strauss, ``Advances in distributed optimization
  using probability collectives,'' \emph{Adv. Complex Syst.}, vol.~9, 2011.

\bibitem{distctrl_lan2010}
Y.~Lan, Z.~Lin, M.~Cao, and G.~Yan, ``A distributed reconfigurable control law
  for escorting and patrolling missions using teams of unicycles,'' in
  \emph{Proc. of IEEE CDC}, 2010, pp. 5456--5461.

\bibitem{shepherding_lien2004}
J.-M. Lien, O.~B. Bayazit, R.~T. Sowell, S.~Rodriguez, and N.~M. Amato,
  ``Shepherding behaviours,'' in \emph{Proc. of IEEE ICRA}, 2004.

\bibitem{distopt_monti2014}
E.~Montijano and A.~R. Mosteo, ``Efficient multi-robot formations using
  distributed optimization,'' in \emph{Proc. of IEEE CDC}, 2014, pp.
  6167--6172.

\bibitem{globalref_monti2014}
E.~Montijano, D.~Zhou, M.~Schwager, and C.~Sagues, ``Distributed formation
  control without a global reference frame,'' in \emph{Proc. of ACC}, 2014, pp.
  3862--3867.

\bibitem{shepherding_strombom2014}
D.~Strömbom, R.~P. Mann, A.~M. Wilson, S.~Hailes, A.~J. Morton, D.~J.~T.
  Sumpter, and A.~J. King, ``Solving the shepherding problem: heuristics for
  herding autonomous, interacting agents.'' \emph{J. R. Soc. Interface},
  vol.~11, no. 100, 2014.

\bibitem{ee_antonelli2007}
G.~Antonelli, F.~Arrichiello, and S.~Chiaverini, ``The entrapment/escorting
  mission for a multi-robot system: Theory and experiments,'' in \emph{Proc. of
  IEEE/ASME AIM}, 2007.

\bibitem{ee_antonelli2008}
------, ``The entrapment/escorting mission,'' \emph{IEEE Robot. Autom. Mag.},
  vol.~15, no.~1, pp. 22--29, 2008.

\bibitem{ee_mas2009}
I.~Mas, S.~Li, J.~Acain, and C.~Kitts, ``Entrapment/escorting and patrolling
  missions in multi-robot cluster space control,'' in \emph{Proc. of IEEE/RSJ
  IROS}, 2009.

\bibitem{Wu_coord_2021}
Y.~Wu, S.~Wu, and X.~Hu, ``Cooperative path planning of {UAV}s \& {UGV}s for a
  persistent surveillance task in urban environments,'' \emph{IEEE Internet
  Things J.}, vol.~8, no.~6, pp. 4906--4919, 2021.

\bibitem{TASE_coord_2019}
S.~G. Manyam, K.~Sundar, and D.~W. Casbeer, ``Cooperative routing for an
  air–ground vehicle team—exact algorithm, transformation method, and
  heuristics,'' \emph{IEEE Trans. Autom. Sci. Eng.}, vol.~17, no.~1, pp.
  537--547, 2020.

\bibitem{mars_nilsson2018}
P.~Nilsson, S.~Haesaert, R.~Thakker, K.~Otsu, C.-I. Vasile, A.-A.
  Agha-Mohammadi, R.~M. Murray, and A.~D. Ames, ``Toward specification-guided
  active {Mars} exploration for cooperative robot teams,'' in \emph{Proc. of
  RSS}, 2018.

\bibitem{mars_sasaki2020}
T.~Sasaki, K.~Otsu, R.~Thakker, S.~Haesaert, and A.-a. Agha-mohammadi, ``Where
  to map? {I}terative rover-copter path planning for {Mars} exploration,''
  \emph{IEEE Robot Autom. Lett.}, vol.~5, no.~2, pp. 2123--2130, 2020.

\bibitem{mars_folsom2021}
L.~Folsom, M.~Ono, K.~Otsu, and H.~Park, ``Scalable information-theoretic path
  planning for a rover-helicopter team in uncertain environments,'' \emph{Int.
  J. Adv. Robot Syst.}, vol.~18, no.~2, 2021.

\bibitem{brian2021}
K.~M.~B. Lee, F.~H. Kong, R.~Cannizzaro, J.~L. Palmer, D.~Johnson, C.~Yoo, and
  R.~Fitch, ``An upper confidence bound for simultaneous exploration and
  exploitation in heterogeneous multi-robot systems,'' in \emph{Proc. of IEEE
  ICRA}, 2021.

\bibitem{atanasov2014}
N.~Atanasov, J.~Le~Ny, K.~Daniilidis, and G.~J. Pappas, ``Information
  acquisition with sensing robots: Algorithms and error bounds,'' in
  \emph{Proc. of IEEE ICRA}, 2014.

\bibitem{atanasov2015}
------, ``Decentralized active information acquisition: Theory and application
  to multi-robot {SLAM},'' in \emph{Proc. of IEEE ICRA}, 2015.

\bibitem{schlotfeldt2018}
B.~Schlotfeldt, D.~Thakur, N.~Atanasov, V.~Kumar, and G.~J. Pappas, ``Anytime
  planning for decentralized multirobot active information gathering,''
  \emph{IEEE Robot Autom. Lett.}, vol.~3, no.~2, pp. 1025--1032, 2018.

\bibitem{jen2021}
J.~Wakulicz, H.~Kong, and S.~Sukkarieh, ``Active information acquisition under
  arbitrary unknown disturbances,'' in \emph{Proc. of IEEE ICRA}, 2021.

\bibitem{krause2007near}
A.~Krause and C.~Guestrin, ``Near-optimal observation selection using
  submodular functions,'' in \emph{Proc. of AAAI}, vol.~7, 2007, pp.
  1650--1654.

\bibitem{sensorschedACC2018}
A.~Hashemi, M.~Ghasemi, H.~Vikalo, and U.~Topcu, ``A randomized greedy
  algorithm for near-optimal sensor scheduling in large-scale sensor
  networks,'' in \emph{Proc. of ACC}, 2018, pp. 1027--1032.

\bibitem{activeprcpt_ghasemi2019}
M.~Ghasemi and U.~Topcu, ``Online active perception for partially observable
  markov decision processes with limited budget,'' in \emph{Proc. of IEEE CDC},
  2019, pp. 6169--6174.

\bibitem{sensorsched2010}
M.~Shamaiah, S.~Banerjee, and H.~Vikalo, ``Greedy sensor selection: Leveraging
  submodularity,'' in \emph{Proc. of IEEE CDC}, 2010, pp. 2572--2577.

\bibitem{chen2020broadcast}
Y.~Chen, L.~Zhao, K.~M.~B. Lee, C.~Yoo, S.~Huang, and R.~Fitch, ``Broadcast
  your weaknesses: cooperative active pose-graph slam for multiple robots,''
  \emph{IEEE Robot. Autom. Lett.}, vol.~5, no.~2, pp. 2200--2207, 2020.

\bibitem{to2021estimation}
K.~C. To, F.~H. Kong, K.~M.~B. Lee, C.~Yoo, S.~Anstee, and R.~Fitch,
  ``Estimation of spatially-correlated ocean currents from ensemble forecasts
  and online measurements,'' in \emph{Proc. of IEEE ICRA}.\hskip 1em plus 0.5em
  minus 0.4em\relax IEEE, 2021, pp. 2301--2307.

\bibitem{lee2018active}
K.~M.~B. Lee, J.~J.~H. Lee, C.~Yoo, B.~Hollings, and R.~Fitch, ``Active
  perception for plume source localisation with underwater gliders,'' in
  \emph{Proc. of ARAA ACRA}, 2018.

\bibitem{lee2019online}
K.~M.~B. Lee, C.~Yoo, B.~Hollings, S.~D. Anstee, S.~Huang, and R.~Fitch,
  ``Online estimation of ocean current from sparse {GPS} data for underwater
  vehicles,'' in \emph{Proc. of IEEE ICRA}, 2019, pp. 3443--3449.

\bibitem{jen2022}
J.~Wakulicz, K.~M.~B. Lee, C.~Yoo, T.~Vidal-Calleja, and R.~Fitch,
  ``Informative planning for worst-case error minimisation in sparse {G}aussian
  process regression,'' in \emph{Proc. of IEEE ICRA}, 2022.

\bibitem{levine2018reinforcement}
S.~Levine, ``Reinforcement learning and control as probabilistic inference:
  Tutorial and review,'' \emph{arXiv preprint arXiv:1805.00909}, 2018.

\bibitem{rstl_brian2021}
K.~M.~B. Lee, C.~Yoo, and R.~Fitch, ``Signal temporal logic synthesis as
  probabilistic inference,'' in \emph{Proc. of IEEE ICRA}, 2021.

\bibitem{yoo2015control}
C.~Yoo and C.~Belta, ``Control with probabilistic signal temporal logic,''
  \emph{arXiv preprint arXiv:1510.08474}, 2015.

\bibitem{yoo2013provably}
C.~Yoo, R.~Fitch, and S.~Sukkarieh, ``Provably-correct stochastic motion
  planning with safety constraints,'' in \emph{Proc. of IEEE ICRA}.\hskip 1em
  plus 0.5em minus 0.4em\relax IEEE, 2013, pp. 981--986.

\bibitem{yoo2016online}
------, ``Online task planning and control for fuel-constrained aerial robots
  in wind fields,'' \emph{Int. J. of Rob. Res.}, vol.~35, no.~5, pp. 438--453,
  2016.

\end{thebibliography}




\end{document}